\documentclass{article}
\usepackage{microtype}
\usepackage{graphicx}
\usepackage{booktabs} 
\usepackage{subcaption}

\usepackage{hyperref}

\usepackage[accepted]{icml2024}

\usepackage{amsmath}
\usepackage{amssymb}
\usepackage{mathtools}
\usepackage{amsthm}
\usepackage{multirow}

\usepackage[capitalize,noabbrev]{cleveref}

\theoremstyle{plain}
\newtheorem{theorem}{Theorem}[section]

\theoremstyle{definition}

\theoremstyle{remark}

\usepackage[textsize=tiny]{todonotes}

\def \ie {\emph{i.e.}}
\def \eg {\emph{e.g.}}
\def \etc {\emph{etc}}

\icmltitlerunning{Decomposition Ascribed Synergistic Learning for Unified Image Restoration}

\date{}
\begin{document}

\twocolumn[
\icmltitle{Decomposition Ascribed Synergistic Learning for Unified Image Restoration}



\icmlsetsymbol{equal}{*}

\begin{icmlauthorlist}
\icmlauthor{Jinghao Zhang}{ustc}
\icmlauthor{Feng Zhao}{ustc}
\end{icmlauthorlist}

\icmlaffiliation{ustc}{Department of Automation,  University of Science and Technology of China, Hefei, China}

\icmlcorrespondingauthor{Jinghao Zhang}{jhaozhang@mail.ustc.edu.cn}
\icmlcorrespondingauthor{Feng Zhao}{fzhao956@ustc.edu.cn}

\icmlkeywords{Machine Learning, ICML}

\vskip 0.3in
]



\printAffiliationsAndNotice{}  

\begin{abstract}
Learning to restore multiple image degradations within a single model is quite beneficial for real-world applications.
Nevertheless, existing works typically concentrate on regarding each degradation independently, while their relationship has been less exploited to ensure the synergistic learning.
To this end, we revisit the diverse degradations through the lens of singular value decomposition, with the observation that the decomposed singular vectors and singular values naturally undertake the different types of degradation information, dividing various restoration tasks into two groups, \ie, singular vector dominated and singular value dominated.
The above analysis renders a more unified perspective to ascribe the diverse degradations, compared to previous task-level independent learning.
The dedicated optimization of degraded singular vectors and singular values inherently utilizes the potential relationship among diverse restoration tasks, attributing to the Decomposition Ascribed Synergistic Learning (DASL).
Specifically, DASL comprises two effective operators, namely, Singular VEctor Operator (SVEO) and Singular VAlue Operator (SVAO), to favor the decomposed optimization, which can be lightly integrated into existing image restoration backbone.
Moreover, the congruous decomposition loss has been devised for auxiliary.
Extensive experiments on blended five image restoration tasks demonstrate the effectiveness of our method.
\end{abstract}

\section{Introduction}
\label{submission}
Image restoration aims to recover the latent clean images from their degraded observations, and has been widely applied to a series of real-world scenarios, such as photo processing, autopilot, and surveillance. 
Compared to single-degradation removal \cite{zhou2021image,xiao2022image,qin2020ffa,song2023vision,lehtinen2018noise2noise,lee2022ap,pan2020cascaded,nah2021ntire,UHDFourICLR2023,zhang2022noiser}, the recent flourished multi-degradation learning methods have gathered considerable attention, due to their convenient deployment.
However, every rose has its thorn. 
How to ensure the synergy among multiple restoration tasks demands a dedicated investigation, and it is imperative to include their implicit relationship into consideration.

\begin{figure*}[t]
\small
\begin{center}
   \includegraphics[width=0.99\linewidth]{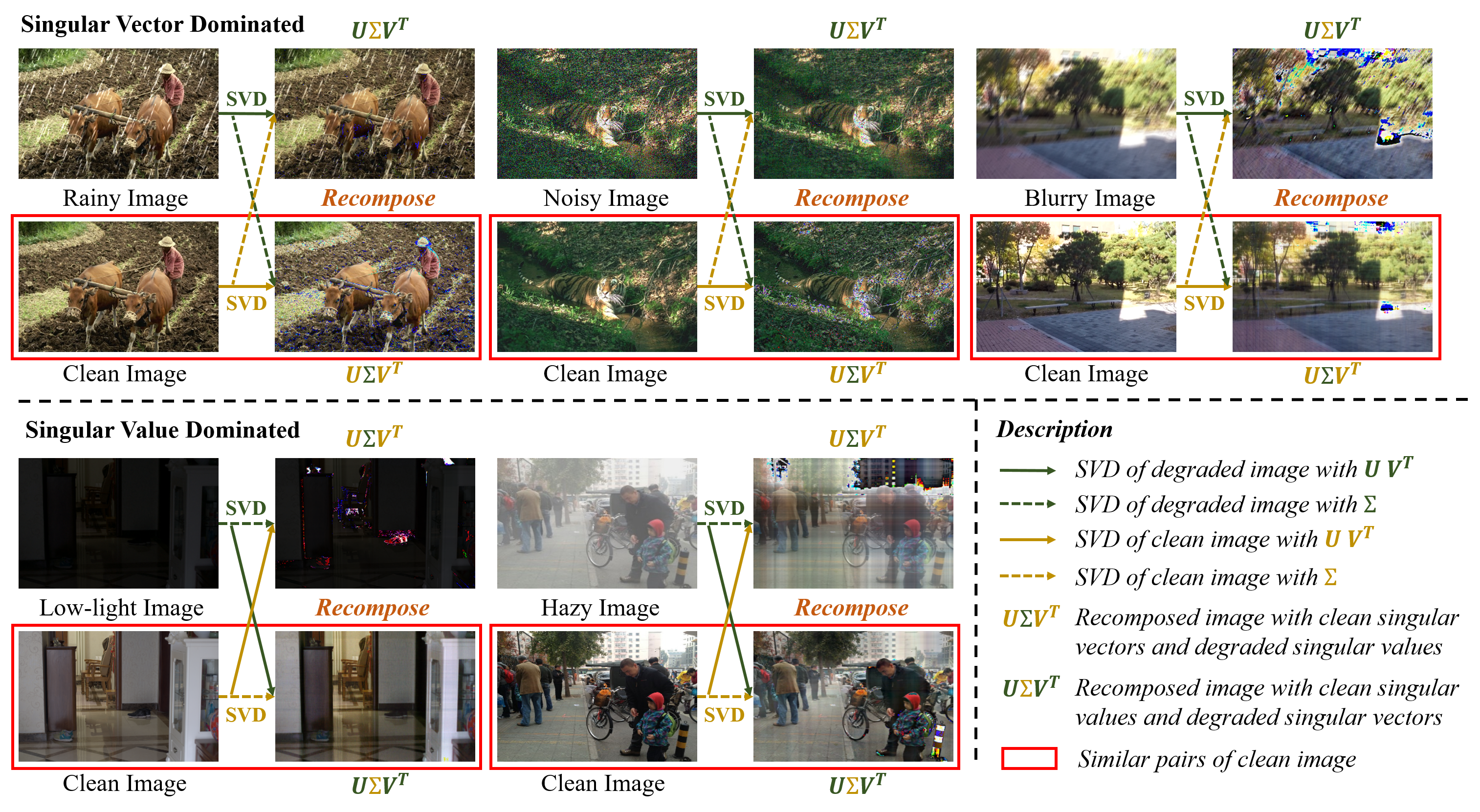}
\end{center}
\vspace{-1em}
    \caption{An illustration of the decomposition ascribed analysis on various image restoration tasks through the lens of the singular value decomposition. The decomposed singular vectors and singular values undertake the different types of degradation information as we recompose the degraded image with portions of the clean counterpart, ascribing diverse restoration tasks into two groups, \ie, singular vector dominated \textit{rain, noise, blur}, and singular value dominated \textit{low-light, haze}. Dedicated to the decomposed optimization of the degraded singular vectors and singular values rendering a more unified perspective for synergistic learning, compared to previous task-level independent learning.}
\label{fig:tease}
\vspace{-1.8em}
\end{figure*}

Generally, existing multi-degradation learning methods concentrated on regarding each degradation independently.
For instance, \cite{chen2021pre,li2020all,valanarasu2022transweather} propose to deal with different restoration tasks through separate subnetworks or distinct transformer queries. \cite{li2022all,chen2022learning} propose to distinguish diverse degradation representations via contrastive learning.
Remarkably, there are also few attempts devoted to duality degradation removal with synergistic learning.
\cite{zhangself} proposes to leverage the blurry and noisy pairs for joint restoration as their inherent complementarity during digital imaging.
\cite{zhou2022lednet} proposes a unified network with low-light enhancement encoder and deblurring decoder to address hybrid distortion.
\cite{wang2022relationship} proposes to quantify the relationship between arbitrary two restoration tasks, and improve the performance of the anchor task with the aid of another task.
However, few efforts have been made toward synergistic learning among more restoration tasks, 
and there is a desperate lack of perspective to revisit the diverse degradations for combing their implicit relationship, which set up the stage for this paper.

\begin{figure*}[t]\small
\begin{center}
   \includegraphics[width=1\linewidth]{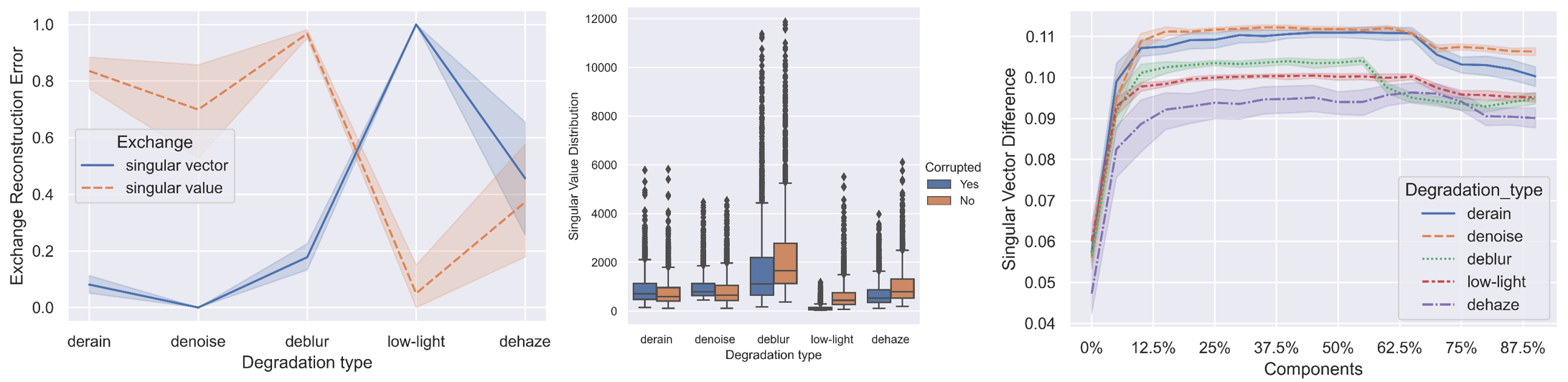}
\end{center}
\vspace{-1em}
   \caption{The statistic validation that the decomposed singular vectors and singular values undertake the different types of degradation information. (a) The reconstruction error between the recomposed image and paired clean image on five common image restoration tasks.
   Low error denotes the degradation primarily distributed in the replaced portion of the image.
   (b) The boxplot comparison of singular value distribution between the degraded image and corresponding clean image, where the singular value dominated \textit{low-light} and \textit{haze} exhibit extraordinary difference. (c) The singular vector difference on separate orders of the component between the degraded image and clean image, where the singular vector dominated \textit{rain}, \textit{noise}, and \textit{blur} present more disparity. The results are obtained under calculation on 100 images for each restoration task.}
\label{fig:stas}
\vspace{-1.5em}
\end{figure*}

To solve the above problem, we propose to revisit diverse degradations through the lens of singular value decomposition, and conduct experiments on five common image restoration tasks, including image deraining, dehazing, denoising, deblurring, and low-light enhancement.
As shown in \cref{fig:tease}, it can be observed that the decomposed singular vectors and singular values naturally undertake the different types of degradation information,
in that the corruptions fade away when we recompose the degraded image with portions of the clean counterpart.
Thus, various restoration tasks can be ascribed into two groups, \ie,  singular vector dominated degradations and singular value dominated deagradations. 
The statistic results in \cref{fig:stas} further validate this phenomenon, where the quantified recomposed image quality and the singular distribution discrepancy have been presented.
Therefore, the potential relationship emerged among diverse restoration tasks could be inherently utilized through the decomposed optimization of singular vectors and singular values, considering their ascribed common properties and significant discrepancies.

In this way, we decently convert the previous task-level independent learning into more unified singular vectors and singular values learning, and form our method, Decomposition Ascribed Synergistic Learning (DASL).
Basically, one straightforward way to implement our idea is to directly perform the decomposition on latent high-dimensional tensors, and conduct the optimization for decomposed singular vectors and singular values, respectively. 
However, the huge computational overhead is non-negligible. 
To this end, two effective operators have been developed to favor the decomposed optimization, namely, Singular VEctor Operator (SVEO) and Singular VAlue Operator (SVAO).
Specifically, SVEO takes advantage of the fact that the orthogonal matrices multiplication makes no effect on singular values and only impacts singular vectors, which can be realized through simple regularized convolution layer.
SVAO resorts to the signal formation homogeneity between Singular Value Decomposition and the Inverse Discrete Fourier Transform, which can both be regarded as a weighted sum on a set of basis. 
While the decomposed singular values and the transformed fourier coefficients inherently undertake the same role for linear combination. 
And the respective base components share similar principle, \ie from outline to details.
Therefore, with approximate derivation, the unattainable singular values optimization can be translated to accessible spectrum maps.
We show that the fast fourier transform is substantially faster than the singular value decomposition. 
Furthermore, the congruous singular decomposition loss has been devised for auxiliary. 
The proposed DASL can be lightly integrated into existing image restoration backbone for decomposed optimization.

The contributions of this work are summarized below:
\begin{itemize}
\item We take a step forward to revisit the diverse degradations through the lens of singular value decomposition, and observe that the decomposed singular vectors and singular values naturally undertake the different types of degradation information, ascribing various restoration tasks into two groups, \ie singular vector dominated and singular value dominated.
\item We propose the Decomposition Ascribed Synergistic Learning (DASL) to dedicate the decomposed optimization of degraded singular vectors and singular values respectively, which inherently utilizes the potential relationship among diverse restoration tasks.
\item Two effective operators have been developed to favor the decomposed optimization, along with a congruous decomposition loss, which can be lightly integrated into existing image restoration backbone. Extensive experiments on five image restoration tasks demonstrate the effectiveness of our method.
\end{itemize}

\section{Related work}

\noindent 
\textbf{Image Restoration.} Image restoration aims to recover the latent clean images from degraded observations, which has been a long-term problem. Traditional image restoration methods typically concentrated on incorporating various natural image priors along with hand-crafted features for specific degradation removal \cite{babacan2008variational,he2010single,kundur1996blind}. 
Recently,  learning-based methods have made compelling progress on various image restoration tasks, including image denoising \cite{lehtinen2018noise2noise,lee2022ap}, image deraining \cite{zhou2021image,xiao2022image}, image deblurring \cite{pan2020cascaded,nah2021ntire}, image dehazing \cite{zheng2021ultra,song2023vision}, and low-light image enhancement \cite{UHDFourICLR2023,guo2020zero}, \etc.
Moreover, numerous general image restoration methods have also been proposed. \cite{zamir2021multi,zamir2022learning,fu2021unfolding} propose the balance between contextual information and spatial details.
\cite{mou2022deep} formulates the image restoration via proximal mapping for iterative optimization.
\cite{zhou2022deep,zhou2023fourmer} proposes to exploit the frequency characteristics to handle diverse degradations.
Additionally, various transformer-based methods \cite{zamir2022restormer,liu2022tape,liang2021swinir,wang2022uformer} have also been investigated, due to their impressive performance in modeling global dependencies and superior adaptability to input contents.

Recently, recovering multiple image degradations within a single model has been coming to the fore, as they are more in line with real-world applications.
\cite{zhangself} proposes to leverage the short-exposure noisy image and the long-exposure blurry image for joint restoration as their inherent complementarity during digital imaging.
\cite{zhou2022lednet} proposes a unified network to address low-light image enhancement and image deblurring.
Furthermore, numerous all-in-one fashion methods \cite{chen2021pre,li2020all,li2022all,valanarasu2022transweather,chen2022learning} have been proposed to deal with multiple degradations.
\cite{zhang2023ingredient} proposes to correlate various degradations through underlying degradation ingredients. While \cite{park2023all} advocates to separate the diverse degradations propcessing with specific attributed discriminative filters. Besides, most of existing methods concentrated on the network architecture design and few attempts have been made toward exploring the synergy among diverse image restoration tasks.

\begin{figure*}[t]
\begin{center}
   \includegraphics[width=1\linewidth]{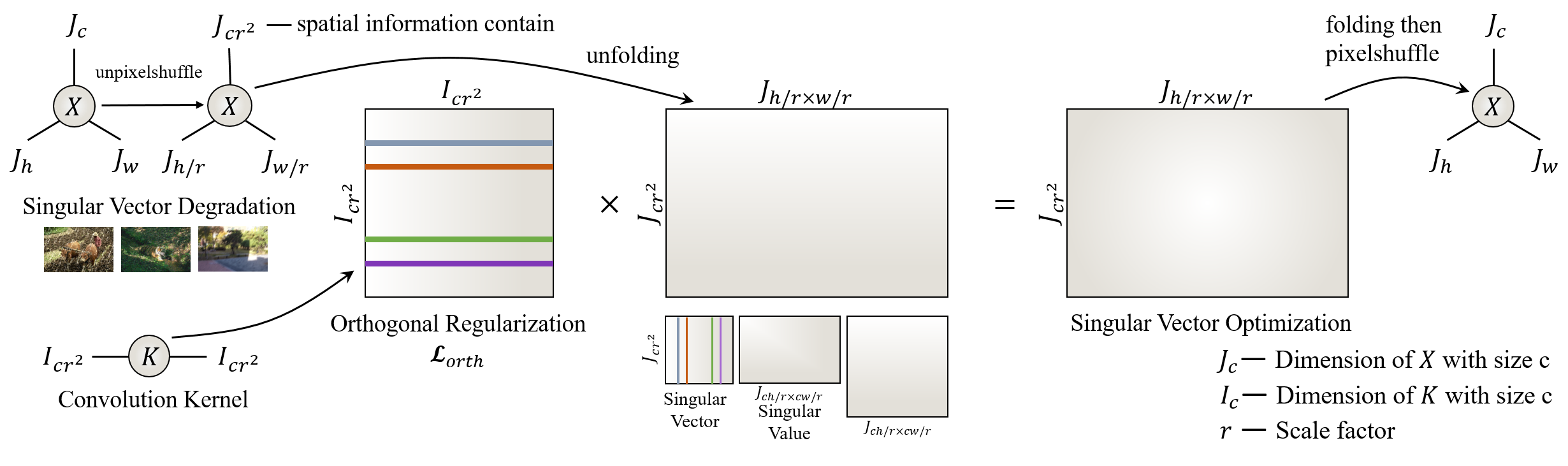}
\end{center}
\vspace{-1.2em}
   \caption{An illustration of the proposed Singular Vector Operator (SVEO), which is dedicated on the optimization of the singular vector dominated degradations, \ie rain, noise, blur. Theorem \ref{theorem:orth} supports the feasibility and the orthogonal regularization $\mathcal{L}_{orth}$ refers to Eq. \ref{eq:orth}.}
\label{fig:sveo}
\vspace{-1em}
\end{figure*}

\noindent
\textbf{Tensor Decomposition.} Tensor decomposition has been widely applied to a series of fields, such as model compression \cite{jie2022fact,obukhov2020t}, neural rendering \cite{obukhov2022tt}, multi-task learning \cite{kanakis2020reparameterizing}, and reinforcement learning \cite{sozykinttopt}.  
In terms of image restoration, a large number of decomposition-based methods have been proposed for hyperspectral and multispectral image restoration \cite{peng2022exact,wang2020hyperspectral,wang2017hyperspectral}, in that establishing the spatial-spectral correlation with low-rank approximation.

Alternatively, a surge of filter decomposition methods toward networks have also been developed.
 \cite{zhang2015accelerating,li2019learning,jaderberg2014speeding} propose to approximate the original filters with efficient representations to reduce the network parameters and inference time.
\cite{kanakis2020reparameterizing} proposes to reparameterize the convolution operators into a non-trainable shared part and several task-specific parts for multi-task learning.
\cite{sun2022singular} proposes to decompose the backbone network and only finetune the singular values to preserve the pre-trained semantic clues for few-shot segmentation.

\section{Method}
In this section, we start with introducing the overall framework of Decomposition Ascribed Synergistic Learning in \cref{sec:overview}, and then elaborate the singular vector operator and singular value operator in \cref{sec:sveo} and \cref{sec:svao}, respectively, which forming our core components.
The optimization objective is briefly presented in \cref{sec:obj}.

\subsection{Overview}
\label{sec:overview}
The intention of the proposed Decomposition Ascribed Synergistic Learning (DASL) is to dedicate the decomposed optimization of degraded singular vectors and singular values respectively, since they naturally undertake the different types of degradation information as observed in \cref{fig:tease,fig:stas}.
And the decomposed optimization renders a more unified perspective to revisit diverse degradations for ascribed synergistic learning.
Through examining the singular vector dominated degradations which containing \textit{rain, noise, blur}, and singular value dominated degradations including \textit{hazy, low-light}, we make the following assumptions: (i) The singular vectors responsible for the content information and spatial details. (ii) The singular values represent the global statistical properties of the image.
Therefore, the optimization of the degraded singular vectors could be performed throughout the backbone network.
And the optimization for the degraded singular values can be condensed to a few of pivotal positions.
Specifically, we substitute half of the convolution layers with SVEO, which are uniformly distributed across the entire network.
While the SVAOs are only performed at the bottleneck layers of the backbone network.
We ensure the compatibility between the optimized singular values and singular vectors through remaining regular layers, and the proposed DASL can be lightly integrated into existing image restoration backbone for decomposed optimization.

\begin{figure*}[t]
\begin{center}
   \includegraphics[width=1\linewidth]{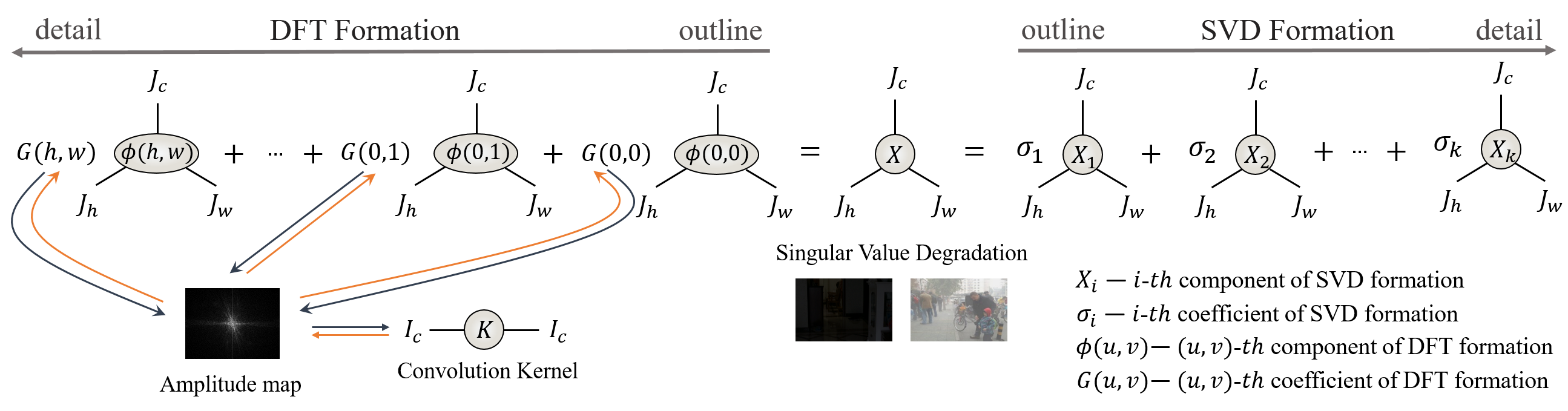}
\end{center}
\vspace{-1.2em}
   \caption{An illustration of the core idea of the proposed Singular Value Operator (SVAO), which is dedicated on the optimization of the singular value dominated degradations, \ie haze and low-light. Two-dimensional signal formations are provided for simplicity.}
\label{fig:svao}
\vspace{-1em}
\end{figure*}

\subsection{Singular Vector Operator}
\label{sec:sveo}
The singular vector operator is proposed to optimize the degraded singular vectors of the latent representation, and supposed to be decoupled with the optimization of singular values. 
Explicitly performing the singular value decomposition on high-dimensional tensors solves this problem naturally with little effort, however, the huge computational overhead is non-negligible. 
Whether can we modify the singular vectors with less computation burden.
The answer is affirmative and lies in the orthogonal matrices multiplication.
\begin{theorem}
For an arbitrary matrix $X \in \mathbb{R}^{h\times w}$ and random orthogonal matrices $P \in \mathbb{R}^{h\times h}, Q\in \mathbb{R}^{w\times w}$, the products of the $PX$, $XQ$, $PXQ$ have the same singular values with the matrix $X$.
\label{theorem:orth}
\end{theorem}
We provide the proof of theorem \ref{theorem:orth} in the Appendix \ref{sec: proof_1}.
In order to construct the orthogonal regularized operator to process the latent representation, the form of the convolution operation is much eligible than matrix multiplication, which is agnostic to the input resolution. 
Hence the distinction between these two forms of operation ought to be taken into consideration.

Prior works \cite{sedghi2018the,jain1989fundamentals} have shown that the convolution operation $y=conv(x)$ with kernel size $k \times k$ can be transformed to linear matrix multiplication $vec(y)=A \, vec(x)$.
Supposing the processed tensors $y,x \in \mathbb{R}^{1\times n\times n}$ for simplicity, the size of the projection matrix $A$ will come to be $n^{2} \times n^{2}$ with doubly block circulant, which is intolerable to enforce the orthogonal regularization, especially for high-resolution inputs.
Another simple way is to employ the $1 \times 1$ convolution with regularized orthogonality, however, 
the singular vectors of the latent representation along the channel dimension will be changed rather than spatial dimension.

Inspired by this point, SVEO proposes to transpose spatial information of the latent representation  $X \in \mathbb{R}^{c\times h\times w}$ to channel dimension with the ordinary unpixelshuffle operation \cite{shi2016real}, resulting in $X^{'} \in \mathbb{R}^{cr^{2}\times h/r\times w/r}$. And then applying the orthogonal regularized $1 \times 1$ convolution $\mathcal{K} \in \mathbb{R}^{cr^{2}\times cr^{2}}$ in this domain, as shown in \cref{fig:sveo}. 
Thereby, the degraded singular vectors can be revised pertinently, and the common properties among various singular vector dominated degradations can be implicitly exploited.
We note that the differences between SVEO and conventional convolution lie in the following: 
(i) The SVEO is more consistent with the matrix multiplication as it eliminates the overlap operation attached to the convolution.
(ii) The weights of SVEO are reduced to matrix instead of tensor, where the orthogonal regularization can be enforced comfortably.
Besides, compared to the matrix multiplication, SVEO further utilizes the channel redundancy and spatial adaptivity within a local $r \times r$ region for conducive information utilization.
The orthogonal regularization is formulated as
\begin{equation}
    \mathcal{L}_{orth} = \Vert WW^T \odot (\textbf{1}-I) \Vert_F^2,
\label{eq:orth}
\end{equation}
where $W$ represents the weight matrix, $\textbf{1}$ denotes a matrix with all elements set to 1, and $I$ denotes the identity matrix. 

\begin{table}[t]
\normalsize
\caption{Time comparison (ms) between SVD and FFT formation for signal representation on high-dimensional tensor, with supposed size $64 \times 128 \times 128$, where the \textit{Decom.} and \textit{Comp.} represent the decomposition and composition.}
\centering
\setlength{\tabcolsep}{1.8mm}{
\begin{tabular}{c|ccc}
\hline
Formation&\textit{Decom.} time & \textit{Comp.}time& Total time\\
\hline
SVD&180.243 &0.143&180.386\\ 
FFT&0.159&0.190&0.349\\
\hline
\end{tabular}}
\vspace{-1.2em}
\label{Tab:svdfft}
\end{table}

\subsection{Singular Value Operator}
\label{sec:svao}
\begin{figure}[t]
\begin{center}
   \includegraphics[width=1\linewidth]{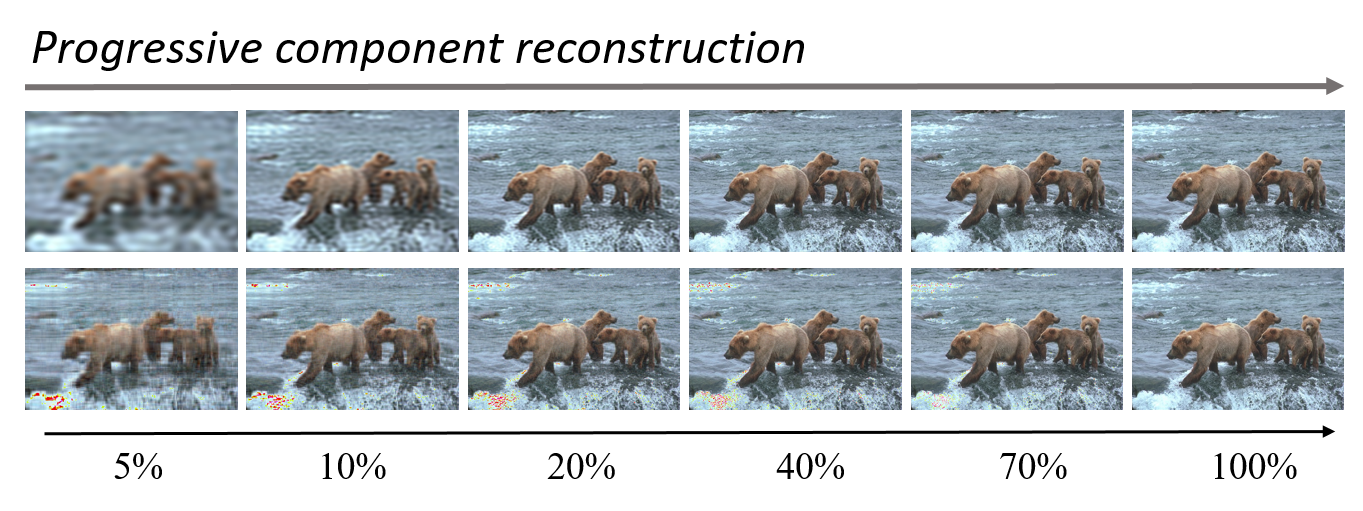}
\end{center}
\vspace{-1.3em}
   \caption{Visual comparison of the progressive reconstruction results with SVD and IDFT components, respectively. First row, IDFT reconstruction result. Second row, SVD reconstruction result. Both conform to the principle from outline to details.}
\label{fig:svddft}
\vspace{-0.8em}
\end{figure}

The singular value operator endeavors to optimize the degraded singular values of the latent representation while supposed to be less entangled with the optimization of singular vectors.
However, considering the inherent inaccessibility of the singular values, it is hard to perform the similar operation as SVEO in the same vein.
To this end, we instead resort to reconnoitering the essence of singular values and found that it is eminently associated with inverse discrete fourier transform.
We provide the formation of a two-dimensional signal represented by singular value decomposition (SVD) and inverse discrete fourier transform (IDFT) in Eq. \ref{eq:svd} and Eq. \ref{eq:fft} as follows.

\begin{equation}
\textstyle
    X = U \Sigma V^{T} = \sum_{i=1}^k \sigma_i u_i v_i^T = \sum_{i=1}^k \sigma_iX_i,
\label{eq:svd}
\end{equation}
where $X \in \mathbb{R}^{h\times w}$ represents the latent representation and $U\in \mathbb{R}^{h\times h}$, $V\in \mathbb{R}^{w\times w}$ represent the decomposed singular vectors with columns $u_i$, $v_i$, $k = min(h,w)$ denotes the rank of $X$. $\Sigma$ represents the singular values with diagonal elements $\sigma_i$.
\begin{align}\label{eq:fft}
    X &= \frac{1}{hw}
\sum_{u=0}^{h-1}\sum_{v=0}^{w-1} G(u,v)e^{j2\pi(\frac{um}{h}+\frac{vn}{w})} \nonumber\\
 &=\frac{1}{hw}\sum_{u=0}^{h-1}\sum_{v=0}^{w-1} G(u,v)\phi(u,v), \; 
\end{align}

where $G(u,v)$ denotes the coefficients of the fourier transform of $X$, and $\phi(u,v)$ denotes the corresponding two-dimensional wave component. $m\in \mathbb{R}^{h-1}$, $n \in\mathbb{R}^{w-1}$.
Observing that both SVD and IDFT formation can be regarded as a weighted sum on a set of basis, \ie $u_iv_i^T$ and $e^{j2\pi(\frac{um}{h}+\frac{vn}{w})}$, while the decomposed singular values $\sigma_i$ and the transformed fourier coefficients $G(u,v)$ inherently undertake the same role for the linear combination of various bases.
In \cref{fig:svddft}, we present the visualized comparison of the reconstruction results using partial components of SVD and IDFT progressively, while both formations conform to the principle from outline to details.
Therefore, we presume that the SVD and IDFT operate in a similar way in terms of signal formation, and the combined coefficients $\sigma_i$ and $G(u,v)$ can be approximated to each other.

In this way, we successfully translate the unattainable singular values optimization to the accessible fourier coefficients optimization, as shown in \cref{fig:svao}.
Considering the decomposed singular values typically characterize the global statistics of the signal, SVAO thus concentrates on the optimization of the norm of $G(u,v)$ for consistency, \ie the amplitude map, since the phase of $G(u,v)$ implicitly represents the structural content \cite{stark2013image,oppenheim1981importance} and more in line with the singular vectors. 
The above two-dimensional signal formation can be easily extended to the three-dimensional tensor to perform the $1 \times 1$ convolution.
Since we adopt the SVAO merely in the bottleneck layers of
the backbone network with low resolution inputs, and the fast fourier transform is substantially faster than the singular value decomposition; see \cref{Tab:svdfft}. The consequent overhead of SVAO can be greatly compressed.
Note that the formation of \cref{eq:fft} is a bit different from the definitive IDFT, and we provide the equivalence proof in the Appendix \ref{sec: proof_2}.

\subsection{Optimization objective}
\label{sec:obj}
The decomposition loss $\mathcal{L}_{dec}$ is developed to favor the decomposed optimization congruously, formulated as
\vspace{-0.6em}
\begin{equation}
    \mathcal{L}_{dec} = \sum_{i=1}^{3} \beta\Vert U_{rec}^{(i)}V_{rec}^{(i)T}-U_{cle}^{(i)}V_{cle}^{(i)T}\Vert_{1} +\Vert\Sigma_{rec}^{(i)}-\Sigma_{cle}^{(i)}\Vert_{1},
\end{equation}
where $U_{cle}$, $V_{cle}$, and $\Sigma_{cle}$ represent the decomposed singular vectors and singular values of the clean image, $U_{rec}$, $V_{rec}$, and $\Sigma_{rec}$ represent the decomposed singular vectors and singular values of the recovered image. For simplicity, we omit the pseudo-identity matrix between $UV^T$ for dimension transformation. $\beta$ denotes the weight.

The overall optimization objective of DASL comprises the orthogonal regularization loss $\mathcal{L}_{orth}$ and the decomposition loss $\mathcal{L}_{dec}$, together with the original loss functions of the integrated backbone network $\mathcal{L}_{ori}$, formulated as
\begin{equation}
    \mathcal{L}_{total} = \mathcal{L}_{ori}+\lambda_{orth}\mathcal{L}_{orth} + \lambda_{dec}\mathcal{L}_{dec} ,
\end{equation}
where $\lambda_{orth}$ and $\lambda_{dec}$ denote the balanced weights.

\newcommand{\baseline}{\color{gray}}
\newcommand{\down}{\color{black}}
\newcommand{\up}{\color{black}}

\section{Experiments}
In this section, we first clarify the experimental settings, and then present the qualitative and quantitative comparison results with eleven baseline methods for unified image restoration. Moreover, extensive ablation experiments are conducted to verify the effectiveness of our method.
\begin{table*}[t]
\scriptsize
\caption{Quantitative results on five common image restoration datasets with state-of-the-art general image restoration and all-in-one methods. The baseline results are in {\baseline grey}.}
\centering
\setlength{\tabcolsep}{2mm}{
\begin{tabular}{lcccccccccc|cc|r}
\toprule[1.3pt]
 &\multicolumn{2}{c}{Rain100L} & \multicolumn{2}{c}{BSD68} & \multicolumn{2}{c}{GoPro}& \multicolumn{2}{c}{SOTS} & \multicolumn{2}{c}{LOL} & \multicolumn{2}{|c|}{\textbf{Average}}&\multirow{2}{*}{\textbf{Params}} \\ 
\textbf{Method} &PSNR$\uparrow$&SSIM$\uparrow$&PSNR$\uparrow$&SSIM$\uparrow$&PSNR$\uparrow$&SSIM$\uparrow$&PSNR$\uparrow$&SSIM$\uparrow$&PSNR$\uparrow$&SSIM$\uparrow$&PSNR$\uparrow$&SSIM$\uparrow$&\\
\midrule
NAFNet&35.56&0.967&31.02&0.883&26.53&0.808&25.23&0.939&20.49&0.809&27.76&0.881&17.11M\\
HINet&35.67&0.969&31.00&0.881&26.12&0.788&24.74&0.937&19.47&0.800&27.40&0.875&88.67M\\
MIRNetV2&33.89&0.954&30.97&0.881&26.30&0.799&24.03&0.927&21.52&0.815&27.34&0.875&5.86M\\
SwinIR&30.78&0.923&30.59&0.868&24.52&0.773&21.50&0.891&17.81&0.723&25.04&0.835&0.91M\\
Restormer&34.81&0.962&31.49&0.884&27.22&0.829&24.09&0.927&20.41&0.806&27.60&0.881&26.13M\\
{\baseline MPRNet}&{\baseline38.16}&{\baseline0.981}&{\baseline31.35}&{\baseline0.889}&{\baseline26.87}&{\baseline0.823}&{\baseline24.27}&{\baseline0.937}&{\baseline20.84}&{\baseline0.824}&{\baseline28.27}&{\baseline0.890}&{\baseline15.74M}\\
{\baseline DGUNet}&{\baseline36.62}&{\baseline0.971}&{\baseline31.10}&{\baseline0.883}&{\baseline27.25}&{\baseline0.837}&{\baseline24.78}&{\baseline0.940}&{\baseline21.87}&{\baseline0.823}&{\baseline28.32}&{\baseline0.891}&{\baseline17.33M}\\
\midrule
DL&21.96&0.762&23.09&0.745&19.86&0.672&20.54&0.826&19.83&0.712&21.05&0.743&2.09M\\
Transweather&29.43&0.905&29.00&0.841&25.12&0.757&21.32&0.885&21.21&0.792&25.22&0.836&37.93M\\
TAPE&29.67&0.904&30.18&0.855&24.47&0.763&22.16&0.861&18.97&0.621&25.09&0.801&1.07M\\
IDR&35.63&0.965&31.60&0.887&27.87&0.846&25.24&0.943&21.34&0.826&28.34&0.893&15.34M\\
{\baseline AirNet}&{\baseline32.98}&{\baseline0.951}&{\baseline30.91}&{\baseline0.882}&{\baseline24.35}&{\baseline0.781}&{\baseline21.04}&{\baseline0.884}&{\baseline18.18}&{\baseline0.735}&{\baseline25.49}&{\baseline0.846}&{\baseline8.93M}\\
\midrule
DASL+MPRNet&38.02&0.980&{\up31.57}&{\up0.890}&{\up26.91}&0.823&{\up25.82}&{\up0.947}&{\up20.96}&{\up0.826}&{\up28.66}&{\up0.893}&{\up15.15M}\\
DASL+DGUNet&{\up36.96}&{\up0.972}&{\up31.23}&{\up0.885}&27.23&0.836&{\up25.33}&{\up0.943}&21.78&0.824&{\up28.51}&{\up0.892}&{\up16.92M}\\
DASL+AirNet&{\up34.93}&{\up0.961}&{\up30.99}&{\up0.883}&{\up26.04}&{\up0.788}&{\up23.64}&{\up0.924}&{\up20.06}&{\up0.805}&{\up27.13}&{\up0.872}&{\up5.41M}\\
\bottomrule[1.5pt]
\end{tabular}}
\vspace{-0.8em}
  \label{Tab:basiline}
\end{table*}

\subsection{Implementation Details}
\noindent
\textbf{Tasks and Metrics.}
We train our method on five image restoration tasks synchronously. The corresponding training set includes Rain200L~\cite{yang2017deep} for image deraining, RESIDE-OTS~\cite{li2018benchmarking} for image dehazing, BSD400~\cite{martin2001database}, WED~\cite{ma2016waterloo} for image denoising, GoPro~\cite{nah2017deep} for image deblurring, and LOL~\cite{Chen2018Retinex} for low-light image enhancement.
For evaluation, 100 image pairs in Rain100L~\cite{yang2017deep}, 500 image pairs in SOTS-Outdoor~\cite{li2018benchmarking}, total 192 images in BSD68~\cite{martin2001database}, Urban100~\cite{huang2015single} and Kodak24~\cite{franzen1999kodak}, 1111 image pairs in GoPro~\cite{nah2017deep}, 15 image pairs in LOL~\cite{Chen2018Retinex} are utilized as the test set.
We report the Peak Signal to Noise Ratio (PSNR) and Structural Similarity (SSIM) as numerical metrics in our experiments.

\noindent
\textbf{Training.} We implement our method on single NVIDIA Geforce RTX 3090 GPU. For fair comparison, all comparison methods have been retrained in the new mixed dataset with their default hyper parameter settings.
We adopt the MPRNet~\cite{zamir2021multi}, DGUNet~\cite{mou2022deep}, and AirNet~\cite{li2022all} as our baseline to validate the proposed Decomposition Ascribed Synergistic Learning.
The entire network is trained with Adam optimizer for 1200 epochs.
We set the batch size as 8 and random crop 128x128 patch from the original image as network input after data augmentation. 
We set the $\beta$ in $\mathcal{L}_{dec}$ as 0.01, and the $\lambda_{orth}$, $\lambda_{dec}$ are set to be 1e-4 and 0.1, respectively.
We perform evaluations every 20 epochs with the highest average PSNR scores as the final parameters result.
More model details and training protocols are presented in the Appendix \ref{sec. model}.

\begin{table}[t]
\vspace{-1.2em}
\centering
\small
\caption{Quantitative results of image denoising on BSD68, Urban100 and Kodak24 datasets (PSNR$\uparrow$).}
\setlength{\tabcolsep}{0.8mm}{
\scriptsize
\begin{tabular}{l|ccc|ccc|ccc}
    \toprule[1.3pt]
     &\multicolumn{3}{c}{BSD68} & \multicolumn{3}{|c}{Urban100} & \multicolumn{3}{|c}{Kodak24}\\ 
    \textbf{Method} &$\sigma$=15&$\sigma$=25&$\sigma$=50&$\sigma$=15&$\sigma$=25&$\sigma$=50&$\sigma$=15&$\sigma$=25&$\sigma$=50\\
    \midrule
    NAFNet&33.67&31.02&27.73&33.14&30.64&27.20&34.27&31.80&28.62\\
    HINet&33.72&31.00&27.63&33.49&30.94&27.32&34.38&31.84&28.52\\
    MIRNetV2&33.66&30.97&27.66&33.30&30.75&27.22&34.29&31.81&28.55\\
    SwinIR&33.31&30.59&27.13&32.79&30.18&26.52&33.89&31.32&27.93\\
    Restormer&34.03&31.49&28.11&33.72&31.26&28.03&34.78&32.37&29.08\\
    {\baseline MPRNet}&{\baseline34.01}&{\baseline31.35}&{\baseline28.08}&{\baseline34.13}&{\baseline31.75}&{\baseline28.41}&{\baseline34.77}&{\baseline32.31}&{\baseline29.11}\\
    {\baseline DGUNet}&{\baseline33.85}&{\baseline31.10}&{\baseline27.92}&{\baseline33.67}&{\baseline31.27}&{\baseline27.94}&{\baseline34.56}&{\baseline32.10}&{\baseline28.91}\\
    \midrule
    DL&23.16&23.09&22.09&21.10&21.28&20.42&22.63&22.66&21.95\\
    Transweather&31.16&29.00&26.08&29.64&27.97&26.08&31.67&29.64&26.74\\
    TAPE&32.86&30.18&26.63&32.19&29.65&25.87&33.24&30.70&27.19\\    IDR&34.11&31.60&28.14&33.82&31.29&28.07&34.78&32.42&29.13\\
    {\baseline AirNet}&{\baseline33.49}&{\baseline30.91}&{\baseline27.66}&{\baseline33.16}&{\baseline30.83}&{\baseline27.45}&{\baseline34.14}&{\baseline31.74}&{\baseline28.59}\\
    \midrule
    DASL+MPRNet&34.16&31.57&28.18&34.21&31.82&28.47&34.91&32.46&29.18\\
    DASL+DGUNet&33.94&31.23&27.94&33.74&31.31&27.96&34.69&32.16&28.93\\
    DASL+AirNet&33.69&30.99&27.68&33.35&30.89&27.46&34.32&31.79&28.61\\
\bottomrule[1.5pt]
\end{tabular}}
\label{Tab:noise}
\end{table}

\begin{table}[t]
\vspace{-1.2em}
    \scriptsize
    \renewcommand\arraystretch {1.01}
    \caption{Comparison of the model size and computation complexity between {\baseline baseline} / DASL.}
    \vspace{-0.2em}
    \centering
    \setlength{\tabcolsep}{2.8mm}{
    \begin{tabular}{l|ccc}
    \toprule[1.3pt]
    Method&Params (M) & FLOPs (B) & Inference Time (s)\\
    \hline
    MPRNet&{\baseline15.74} / 15.15 &{\baseline5575.32} / 2905.14& {\baseline0.241} / 0.210 \\ 
    DGUNet&{\baseline17.33} / 16.92&{\baseline3463.66} / 3020.22&{\baseline0.397} / 0.391\\
    AirNet&{\baseline8.93} / 5.41& {\baseline1205.09} / 767.89 & {\baseline0.459} / 0.190\\
    \bottomrule[1.3pt]
\end{tabular}}
\vspace{-1.2em}
  \label{Tab:overhead}
\end{table}

\subsection{Comparison with state-of-the-art methods}
We compare our DASL with eleven state-of-the-art methods, including general image restoration methods: NAFNet~\cite{chen2022simple}, HINet~\cite{chen2021hinet}, MPRNet~\cite{zamir2021multi}, DGUNet~\cite{mou2022deep}, MIRNetV2~\cite{zamir2022learning}, SwinIR~\cite{liang2021swinir}, Restormer~\cite{zamir2022restormer}, and all-in-one fashion methods: DL~\cite{fan2019general}, Transweather~\cite{valanarasu2022transweather}, TAPE~\cite{liu2022tape}, AirNet~\cite{li2022all} and IDR~\cite{zhang2023ingredient} on five common image restoration tasks.

Table \ref{Tab:basiline} reports the quantitative comparison results.
It can be observed that the performance of the general image restoration methods is systematically superior to the professional all-in-one fashion methods when more degradations are involved, attributed to the large model size. 
While our DASL further advances the backbone network capability with fewer parameters, owing to the implicit synergistic learning.
Consistent with existing unified image restoration methods~\cite{zamir2022restormer,li2022all}, we report the detailed denoising results at different noise ratio in \cref{Tab:noise}, where the performance gain are consistent.

In ~\cref{Tab:overhead}, we present the computation overhead involved in DASL, where the FLOPs and inference time are calculated over 100 testing images with the size of 512×512. 
It can be observed that our DASL substantially reduces the computation complexity of the baseline methods with considerable inference acceleration, \eg 12.86\% accelerated on MPRNet and 58.61\% accelerated on AirNet.
We present the bountiful visual comparison results in the Appendix \ref{sec:visual}, while our DASL exhibits superior visual recovery quality, \ie more precise details in singular vector dominated degradations and more stable global recovery in singular value dominated degradations.

\begin{table*}[t]
\renewcommand\arraystretch {1.1}
\caption{Ablation experiments on the components design.}
\centering
\scriptsize
\setlength{\tabcolsep}{1.05mm}{
\begin{tabular}{l|cccc|cccccccccc|cc}
\toprule[1.3pt]
&&&&&\multicolumn{2}{c}{Rain100L} & \multicolumn{2}{c}{BSD68} & \multicolumn{2}{c}{GoPro}& \multicolumn{2}{c}{SOTS} & \multicolumn{2}{c}{LOL}& \multicolumn{2}{c}{\textbf{Avg.}}\\
Method&\textit{SVEO} & \textit{SVAO}  &$\mathcal{L}_{orth}$ & $\mathcal{L}_{dec}$ & PSNR$\uparrow$ & SSIM$\uparrow$ & PSNR$\uparrow$ & SSIM$\uparrow$ & PSNR$\uparrow$ & SSIM$\uparrow$ & PSNR$\uparrow$ & SSIM$\uparrow$ & PSNR$\uparrow$ & SSIM$\uparrow$ & PSNR$\uparrow$ & SSIM$\uparrow$ \\
\hline
{\baseline Baseline}&&&&&{\baseline38.16}&{\baseline0.981}&{\baseline31.35}&{\baseline0.889}&{\baseline26.87}&{\baseline0.823}&{\baseline24.27}&{\baseline0.937}&{\baseline20.84}&{\baseline0.824}&{\baseline28.27}&{\baseline0.890} \\ 
\hline
With no orth. \textit{SVEO}&\checkmark&&&&37.73&0.981&31.31&0.889&26.79&0.819&24.63& 0.939&20.83&0.824&28.26&0.890 \\
With \textit{SVAO}&&\checkmark&&&37.92&0.980&31.41&0.889&26.85&0.821&25.58&0.943&21.05&0.828&28.56&0.892 \\
With \textit{SVEO}&\checkmark&&\checkmark&&38.04&0.981&31.46&0.890&26.97&0.826&25.53&0.945&20.76&0.822&28.55&0.893 \\
With \textit{SVEO} and \textit{SVAO}&\checkmark&\checkmark&\checkmark&&38.01&0.980&31.53&0.890&26.94&0.825&25.63&0.948&20.92&0.826&28.61&0.893 \\
With $\mathcal{L}_{dec}$ &&&&\checkmark&\textbf\underline38.10&0.982&31.39&0.889&26.78&0.819&24.70&0.942&20.98&0.827&28.39&0.892\\
\hline
DASL+MPRNet&\checkmark&\checkmark&\checkmark&\checkmark&38.02&0.980&31.57&0.890&26.91&0.823&25.82&0.947&20.96&0.826&28.66&0.893\\
\bottomrule[1.5pt]
\end{tabular}}
\vspace{-1.6em}
  \label{Tab:AbS1}
\end{table*}

\subsection{Abalation Studies}
We present the ablation experiments on the combined degradation dataset with MPRNet as the backbone to verify the effectiveness of our method.
In \cref{Tab:AbS1}, we quantitatively evaluate the two developed operators SVEO and SVAO, and the decomposition loss. The metrics are reported on the each of degradations in detail, from which we can make the following observations:
\textbf{a)} Both SVEO and SVAO are beneficial for advancing the unified degradation restoration performance, attributing to the ascribed synergistic learning.
\textbf{b)} The congruous decomposition loss is capable to work alone, and well collaborated with developed operators for decomposed optimization.
\textbf{c)} The orthogonal regularization is crucial to the reliable optimization of SVEO for preventing the performance drop.

To further verify the scalability of the decomposed optimization, \cref{Tab:decomopti} evaluates the performance with partially trained on singular vector dominated degradations (\textit{vec}.) and singular value dominated degradations (\textit{val.}). 
While some properties have been observed: 
\textbf{a)} Basically, the baseline methods concentrate on the trainable degradations, while our DASL further contemplates the untrainable ones in virtue of its slight task dependency. 
\textbf{b)} The performance of MPRNet on \textit{val.} is unattainable due to the non-convergence, however, our DASL successfully circumvents this drawback owing to the more unified decomposed optimization on singular values rather than task-level learning.
\textbf{c)} The \textit{vec.} seems to be supportive to the restoration performance of \textit{val.}, see the comparison of \cref{Tab:basiline,Tab:decomopti}, indicating the potential relationship among decomposed two types of degradations.

\begin{figure}[t]
\vspace{-0.4em}
\begin{center}
   \includegraphics[width=1\linewidth]{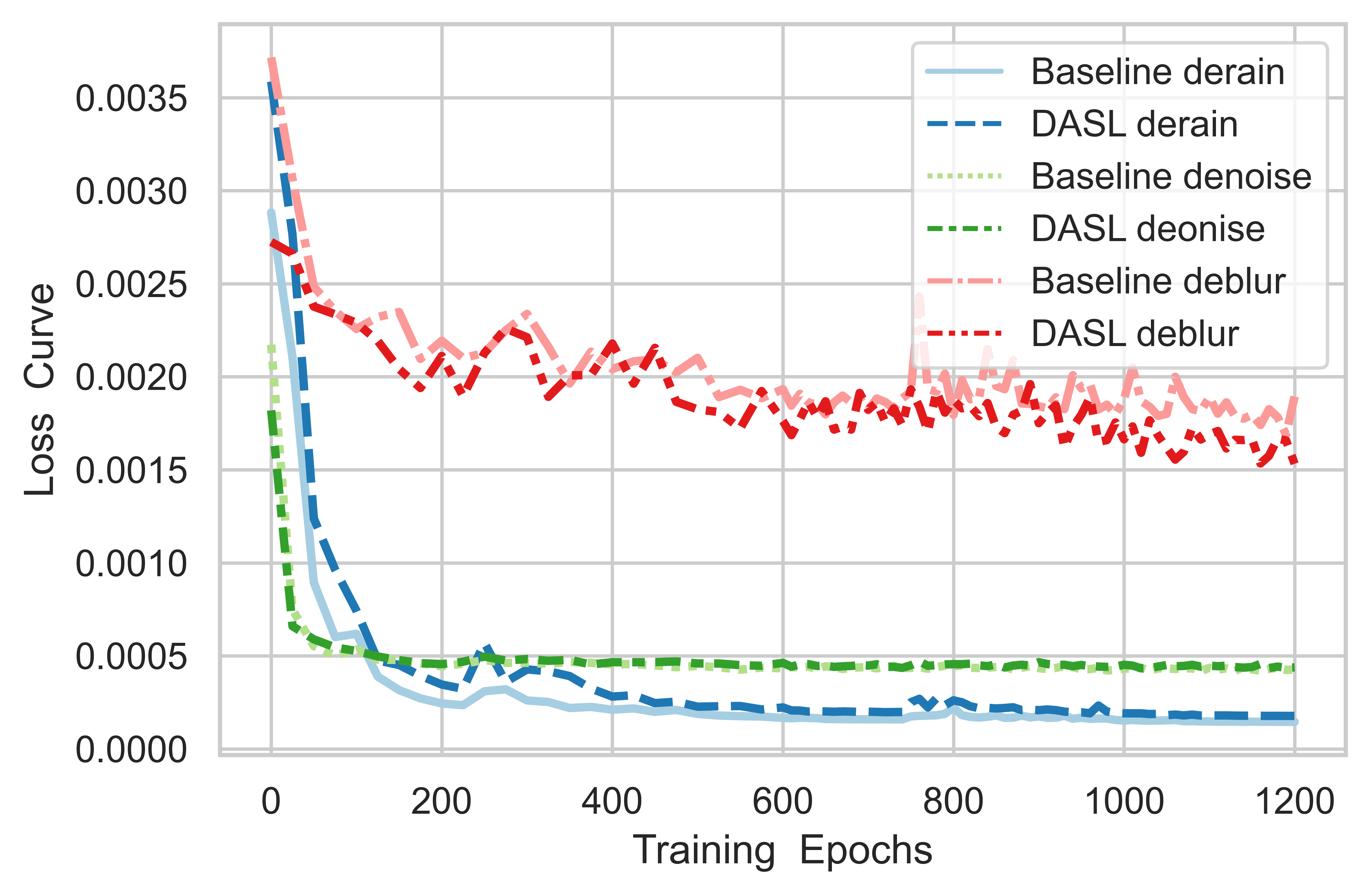}
\end{center}
\vspace{-2em}
   \caption{Evaluating the synergy effect through training trajectory between baseline and DASL on \textit{vec}. dominated degradations.}
\label{fig:SVETraing}
\vspace{-1.4em}
\end{figure}

\begin{figure}[t]
\begin{center}
   \includegraphics[width=1\linewidth]{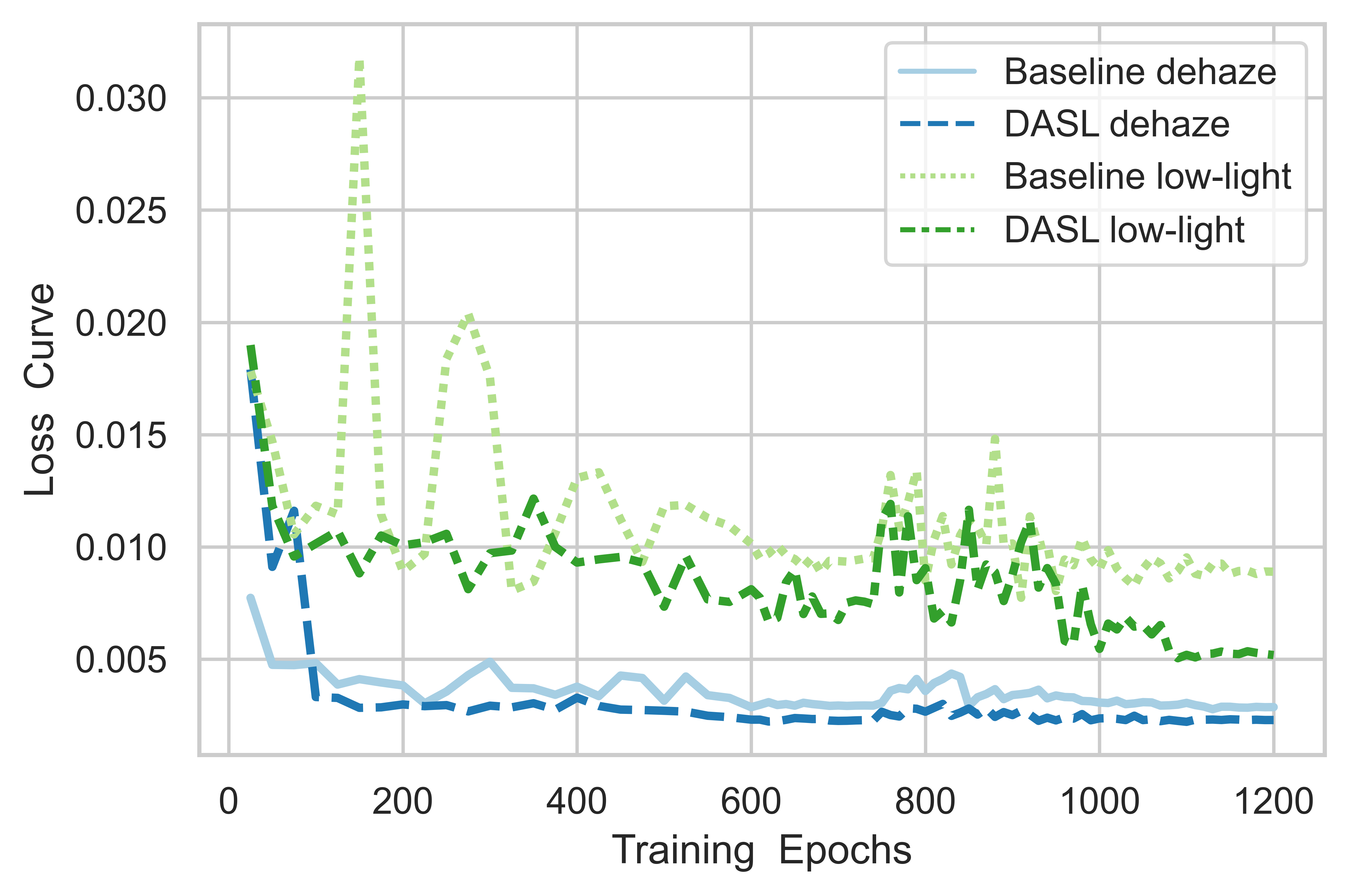}
\end{center}
\vspace{-2em}
   \caption{Evaluating the synergy effect through training trajectory between baseline and DASL on \textit{val}. dominated degradations.}
\label{fig:SVATraing}
\vspace{-1.7em}
\end{figure}

We present the comparison of the training trajectory between baseline and DASL on singular vector dominated and singular value dominated degradations in \cref{fig:SVETraing,fig:SVATraing}.
It can be observed that our DASL significantly suppresses the drastic optimization process, retaining the overall steady to better convergence point with even fewer parameters, attributing to the ascribed synergistic learning.

\subsection{Limitation and Future works}
Despite the great progress that DASL has been made in ascribing the implicit relationship among diverse degradations for synergistic learning, the more sophisticated correlations demand further investigation.
We notice that beyond the decomposed singular vectors and singular values, the distribution discrepancy of various degradations on separate orders of decomposed components exhibits the splendid potential, which can be glimpsed from \cref{fig:stas}.
Additionally, how to leverage and represent their sophisticated relationship in a unified framework remains an open problem.

\begin{table}[t]
\vspace{-1.2em}
\centering
\small
\renewcommand\arraystretch {1.1}
\caption{Evaluating the scalability of decomposed optimization on the full test set with partially trained on singular vector dominated degradations (\textit{vec}.) and singular value dominated degradations (\textit{val}.) (PSNR$\uparrow$).}
\setlength{\tabcolsep}{1.2mm}{
\scriptsize
\begin{tabular}{l|ccccc|c}
    \toprule[1.3pt]
    Method&Rain100L & BSD68 & GoPro &SOTS & LOL & \textbf{Avg.}\\
    \hline
    {\baseline MPRNet (\textit{vec.})}&{\baseline39.47} &{\baseline31.50}&{\baseline27.61}&{\baseline15.91}&{\baseline7.77}&{\baseline24.45} \\ 
    {\baseline MPRNet (\textit{val.})}&{\baseline-}&{\baseline-}&{\baseline-}&{\baseline-}&{\baseline-}&{\baseline-}\\
    {\baseline DGUNet (\textit{vec.})}&{\baseline39.04} &{\baseline31.46}&{\baseline28.22}&{\baseline15.92}&{\baseline7.76}&{\baseline24.48} \\ 
    {\baseline DGUNet (\textit{val.})}&{\baseline23.10}&{\baseline20.39}&{\baseline21.84}&{\baseline24.59}&{\baseline20.45}&{\baseline22.07}\\
    {\baseline AirNet (\textit{vec.})}&{\baseline36.62} &{\baseline31.33}&{\baseline26.35}&{\baseline15.90}&{\baseline7.75}&{\baseline23.59} \\ 
    {\baseline AirNet (\textit{val.})}&{\baseline19.52}&{\baseline19.17}&{\baseline14.47}&{\baseline20.63}&{\baseline16.01}&{\baseline17.96}\\
    \hline
    DASL+MPRNet (\textit{vec.})&39.39&31.63&27.57&17.21&11.23&25.41\\
    DASL+MPRNet (\textit{val.})&21.87&19.96&21.35&25.13&20.33&21.73\\
    DASL+DGUNet (\textit{vec.})&39.11&31.55&28.16&16.87&10.21&25.18\\
    DASL+DGUNet (\textit{val.})&23.19&20.28&22.69&25.05&20.87&22.42\\
    DASL+AirNet (\textit{vec.})&39.11&31.55&28.16&16.87&10.21&25.18\\
    DASL+AirNet (\textit{val.})&23.19&20.28&22.69&25.05&20.87&22.42\\
    \bottomrule[1.5pt]
\end{tabular}}
\label{Tab:decomopti}
\vspace{-1.8em}
\end{table}

\section{Conclusion}
In this paper, we revisited the diverse degradations through the lens of singular value decomposition and observed that the decomposed singular vectors and singular values naturally undertake the different types of degradation information, ascribing various restoration tasks into two groups, \ie singular vector dominated degradations and singular value dominated degradations.
The proposed Decomposition Ascribed Synergistic Learning dedicates the decomposed optimization of degraded singular vectors and singular values respectively, rendering a more unified perspective to inherently utilize the potential relationship among diverse restoration tasks for ascribed synergistic learning.
Furthermore, two effective operators SVEO and SVAO have been developed to favor the decomposed optimization, along with a congruous decomposition loss, which can be lightly integrated into existing image restoration backbone.
Extensive experiments on bunch of image restoration tasks validated the effectiveness of the proposed method and the generality of the SVD-based degradation analysis.

\noindent
\textbf{Broader impact.}
This work potentially release the redundant model deployments in real world scenarios, and sincerely benefits a lot of edge applications with limited resources, such as mobile photography and 24/7 surveillance. 
The privacy of our method may raise potential concerns when considering the removal of some important occlusions in the original images, resulting in the disclosure of private information. Therefore, how to ensure the user-agnostic security of our method needs further investment.

\nocite{langley00}

\bibliography{example_paper}

\begin{thebibliography}{67}
\providecommand{\natexlab}[1]{#1}
\providecommand{\url}[1]{\texttt{#1}}
\expandafter\ifx\csname urlstyle\endcsname\relax
  \providecommand{\doi}[1]{doi: #1}\else
  \providecommand{\doi}{doi: \begingroup \urlstyle{rm}\Url}\fi

\bibitem[Babacan et~al.(2008)Babacan, Molina, and Katsaggelos]{babacan2008variational}
Babacan, S.~D., Molina, R., and Katsaggelos, A.~K.
\newblock Variational bayesian blind deconvolution using a total variation prior.
\newblock \emph{IEEE Transactions on Image Processing}, 18\penalty0 (1):\penalty0 12--26, 2008.

\bibitem[Chen et~al.(2021{\natexlab{a}})Chen, Wang, Guo, Xu, Deng, Liu, Ma, Xu, Xu, and Gao]{chen2021pre}
Chen, H., Wang, Y., Guo, T., Xu, C., Deng, Y., Liu, Z., Ma, S., Xu, C., Xu, C., and Gao, W.
\newblock Pre-trained image processing transformer.
\newblock In \emph{Proceedings of the IEEE/CVF Conference on Computer Vision and Pattern Recognition}, pp.\  12299--12310, 2021{\natexlab{a}}.

\bibitem[Chen et~al.(2021{\natexlab{b}})Chen, Lu, Zhang, Chu, and Chen]{chen2021hinet}
Chen, L., Lu, X., Zhang, J., Chu, X., and Chen, C.
\newblock Hinet: Half instance normalization network for image restoration.
\newblock In \emph{Proceedings of the IEEE/CVF Conference on Computer Vision and Pattern Recognition}, pp.\  182--192, 2021{\natexlab{b}}.

\bibitem[Chen et~al.(2022{\natexlab{a}})Chen, Chu, Zhang, and Sun]{chen2022simple}
Chen, L., Chu, X., Zhang, X., and Sun, J.
\newblock Simple baselines for image restoration.
\newblock In \emph{European Conference on Computer Vision}, pp.\  17--33. Springer, 2022{\natexlab{a}}.

\bibitem[Chen et~al.(2018)Chen, Wenjing, Wenhan, and Jiaying]{Chen2018Retinex}
Chen, W., Wenjing, W., Wenhan, Y., and Jiaying, L.
\newblock Deep retinex decomposition for low-light enhancement.
\newblock In \emph{British Machine Vision Conference}. British Machine Vision Association, 2018.

\bibitem[Chen et~al.(2022{\natexlab{b}})Chen, Huang, Tsai, Yang, Ding, and Kuo]{chen2022learning}
Chen, W.-T., Huang, Z.-K., Tsai, C.-C., Yang, H.-H., Ding, J.-J., and Kuo, S.-Y.
\newblock Learning multiple adverse weather removal via two-stage knowledge learning and multi-contrastive regularization: Toward a unified model.
\newblock In \emph{Proceedings of the IEEE/CVF Conference on Computer Vision and Pattern Recognition}, pp.\  17653--17662, 2022{\natexlab{b}}.

\bibitem[Fan et~al.(2019)Fan, Chen, Yuan, Hua, Yu, and Chen]{fan2019general}
Fan, Q., Chen, D., Yuan, L., Hua, G., Yu, N., and Chen, B.
\newblock A general decoupled learning framework for parameterized image operators.
\newblock \emph{IEEE transactions on pattern analysis and machine intelligence}, 43\penalty0 (1):\penalty0 33--47, 2019.

\bibitem[Franzen(1999)]{franzen1999kodak}
Franzen, R.
\newblock Kodak lossless true color image suite.
\newblock \emph{source: http://r0k. us/graphics/kodak}, 4\penalty0 (2), 1999.

\bibitem[Fu et~al.(2021)Fu, Xiao, Yang, Liu, Xiong, et~al.]{fu2021unfolding}
Fu, X., Xiao, Z., Yang, G., Liu, A., Xiong, Z., et~al.
\newblock Unfolding taylor's approximations for image restoration.
\newblock \emph{Advances in Neural Information Processing Systems}, 34:\penalty0 18997--19009, 2021.

\bibitem[Guo et~al.(2020)Guo, Li, Guo, Loy, Hou, Kwong, and Cong]{guo2020zero}
Guo, C., Li, C., Guo, J., Loy, C.~C., Hou, J., Kwong, S., and Cong, R.
\newblock Zero-reference deep curve estimation for low-light image enhancement.
\newblock In \emph{Proceedings of the IEEE/CVF Conference on Computer Vision and Pattern Recognition}, pp.\  1780--1789, 2020.

\bibitem[He et~al.(2010)He, Sun, and Tang]{he2010single}
He, K., Sun, J., and Tang, X.
\newblock Single image haze removal using dark channel prior.
\newblock \emph{IEEE transactions on pattern analysis and machine intelligence}, 33\penalty0 (12):\penalty0 2341--2353, 2010.

\bibitem[Huang et~al.(2015)Huang, Singh, and Ahuja]{huang2015single}
Huang, J.-B., Singh, A., and Ahuja, N.
\newblock Single image super-resolution from transformed self-exemplars.
\newblock In \emph{Proceedings of the IEEE conference on computer vision and pattern recognition}, pp.\  5197--5206, 2015.

\bibitem[Jaderberg et~al.(2014)Jaderberg, Vedaldi, and Zisserman]{jaderberg2014speeding}
Jaderberg, M., Vedaldi, A., and Zisserman, A.
\newblock Speeding up convolutional neural networks with low rank expansions.
\newblock \emph{arXiv preprint arXiv:1405.3866}, 2014.

\bibitem[Jain(1989)]{jain1989fundamentals}
Jain, A.~K.
\newblock \emph{Fundamentals of digital image processing}.
\newblock Prentice-Hall, Inc., 1989.

\bibitem[Jie \& Deng(2022)Jie and Deng]{jie2022fact}
Jie, S. and Deng, Z.-H.
\newblock Fact: Factor-tuning for lightweight adaptation on vision transformer.
\newblock \emph{arXiv preprint arXiv:2212.03145}, 2022.

\bibitem[Kanakis et~al.(2020)Kanakis, Bruggemann, Saha, Georgoulis, Obukhov, and Van~Gool]{kanakis2020reparameterizing}
Kanakis, M., Bruggemann, D., Saha, S., Georgoulis, S., Obukhov, A., and Van~Gool, L.
\newblock Reparameterizing convolutions for incremental multi-task learning without task interference.
\newblock In \emph{Computer Vision--ECCV 2020: 16th European Conference}, pp.\  689--707. Springer, 2020.

\bibitem[Kundur \& Hatzinakos(1996)Kundur and Hatzinakos]{kundur1996blind}
Kundur, D. and Hatzinakos, D.
\newblock Blind image deconvolution.
\newblock \emph{IEEE signal processing magazine}, 13\penalty0 (3):\penalty0 43--64, 1996.

\bibitem[Lee et~al.(2022)Lee, Son, and Lee]{lee2022ap}
Lee, W., Son, S., and Lee, K.~M.
\newblock Ap-bsn: Self-supervised denoising for real-world images via asymmetric pd and blind-spot network.
\newblock In \emph{Proceedings of the IEEE/CVF Conference on Computer Vision and Pattern Recognition}, pp.\  17725--17734, 2022.

\bibitem[Lehtinen et~al.(2018)Lehtinen, Munkberg, Hasselgren, Laine, Karras, Aittala, and Aila]{lehtinen2018noise2noise}
Lehtinen, J., Munkberg, J., Hasselgren, J., Laine, S., Karras, T., Aittala, M., and Aila, T.
\newblock Noise2noise: Learning image restoration without clean data.
\newblock In \emph{International Conference on Machine Learning}, pp.\  2965--2974. PMLR, 2018.

\bibitem[Li et~al.(2018)Li, Ren, Fu, Tao, Feng, Zeng, and Wang]{li2018benchmarking}
Li, B., Ren, W., Fu, D., Tao, D., Feng, D., Zeng, W., and Wang, Z.
\newblock Benchmarking single-image dehazing and beyond.
\newblock \emph{IEEE Transactions on Image Processing}, 28\penalty0 (1):\penalty0 492--505, 2018.

\bibitem[Li et~al.(2022)Li, Liu, Hu, Wu, Lv, and Peng]{li2022all}
Li, B., Liu, X., Hu, P., Wu, Z., Lv, J., and Peng, X.
\newblock All-in-one image restoration for unknown corruption.
\newblock In \emph{Proceedings of the IEEE/CVF Conference on Computer Vision and Pattern Recognition}, pp.\  17452--17462, 2022.

\bibitem[Li et~al.(2023)Li, Guo, Zhou, Liang, Zhou, Feng, and Loy]{UHDFourICLR2023}
Li, C., Guo, C.-L., Zhou, M., Liang, Z., Zhou, S., Feng, R., and Loy, C.~C.
\newblock Embeddingfourier for ultra-high-definition low-light image enhancement.
\newblock In \emph{ICLR}, 2023.

\bibitem[Li et~al.(2020)Li, Tan, and Cheong]{li2020all}
Li, R., Tan, R.~T., and Cheong, L.-F.
\newblock All in one bad weather removal using architectural search.
\newblock In \emph{Proceedings of the IEEE/CVF conference on computer vision and pattern recognition}, pp.\  3175--3185, 2020.

\bibitem[Li et~al.(2019)Li, Gu, Gool, and Timofte]{li2019learning}
Li, Y., Gu, S., Gool, L.~V., and Timofte, R.
\newblock Learning filter basis for convolutional neural network compression.
\newblock In \emph{Proceedings of the IEEE/CVF International Conference on Computer Vision}, pp.\  5623--5632, 2019.

\bibitem[Liang et~al.(2021)Liang, Cao, Sun, Zhang, Van~Gool, and Timofte]{liang2021swinir}
Liang, J., Cao, J., Sun, G., Zhang, K., Van~Gool, L., and Timofte, R.
\newblock Swinir: Image restoration using swin transformer.
\newblock In \emph{Proceedings of the IEEE/CVF International Conference on Computer Vision}, pp.\  1833--1844, 2021.

\bibitem[Liu et~al.(2022)Liu, Xie, Zhang, Yuan, Chen, Zhou, Li, and Tian]{liu2022tape}
Liu, L., Xie, L., Zhang, X., Yuan, S., Chen, X., Zhou, W., Li, H., and Tian, Q.
\newblock Tape: Task-agnostic prior embedding for image restoration.
\newblock In \emph{European Conference on Computer Vision}, pp.\  447--464. Springer, 2022.

\bibitem[Ma et~al.(2016)Ma, Duanmu, Wu, Wang, Yong, Li, and Zhang]{ma2016waterloo}
Ma, K., Duanmu, Z., Wu, Q., Wang, Z., Yong, H., Li, H., and Zhang, L.
\newblock Waterloo exploration database: New challenges for image quality assessment models.
\newblock \emph{IEEE Transactions on Image Processing}, 26\penalty0 (2):\penalty0 1004--1016, 2016.

\bibitem[Martin et~al.(2001)Martin, Fowlkes, Tal, and Malik]{martin2001database}
Martin, D., Fowlkes, C., Tal, D., and Malik, J.
\newblock A database of human segmented natural images and its application to evaluating segmentation algorithms and measuring ecological statistics.
\newblock In \emph{Proceedings Eighth IEEE International Conference on Computer Vision. ICCV 2001}, volume~2, pp.\  416--423. IEEE, 2001.

\bibitem[Mou et~al.(2022)Mou, Wang, and Zhang]{mou2022deep}
Mou, C., Wang, Q., and Zhang, J.
\newblock Deep generalized unfolding networks for image restoration.
\newblock In \emph{Proceedings of the IEEE/CVF Conference on Computer Vision and Pattern Recognition}, pp.\  17399--17410, 2022.

\bibitem[Nah et~al.(2017)Nah, Hyun~Kim, and Mu~Lee]{nah2017deep}
Nah, S., Hyun~Kim, T., and Mu~Lee, K.
\newblock Deep multi-scale convolutional neural network for dynamic scene deblurring.
\newblock In \emph{Proceedings of the IEEE conference on computer vision and pattern recognition}, pp.\  3883--3891, 2017.

\bibitem[Nah et~al.(2021)Nah, Son, Lee, Timofte, and Lee]{nah2021ntire}
Nah, S., Son, S., Lee, S., Timofte, R., and Lee, K.~M.
\newblock Ntire 2021 challenge on image deblurring.
\newblock In \emph{Proceedings of the IEEE/CVF Conference on Computer Vision and Pattern Recognition}, pp.\  149--165, 2021.

\bibitem[Obukhov et~al.(2020)Obukhov, Rakhuba, Georgoulis, Kanakis, Dai, and Van~Gool]{obukhov2020t}
Obukhov, A., Rakhuba, M., Georgoulis, S., Kanakis, M., Dai, D., and Van~Gool, L.
\newblock T-basis: a compact representation for neural networks.
\newblock In \emph{International Conference on Machine Learning}, pp.\  7392--7404. PMLR, 2020.

\bibitem[Obukhov et~al.(2022)Obukhov, Usvyatsov, Sakaridis, Schindler, and Van~Gool]{obukhov2022tt}
Obukhov, A., Usvyatsov, M., Sakaridis, C., Schindler, K., and Van~Gool, L.
\newblock Tt-nf: Tensor train neural fields.
\newblock \emph{arXiv preprint arXiv:2209.15529}, 2022.

\bibitem[Oppenheim \& Lim(1981)Oppenheim and Lim]{oppenheim1981importance}
Oppenheim, A.~V. and Lim, J.~S.
\newblock The importance of phase in signals.
\newblock \emph{Proceedings of the IEEE}, 69\penalty0 (5):\penalty0 529--541, 1981.

\bibitem[Pan et~al.(2020)Pan, Bai, and Tang]{pan2020cascaded}
Pan, J., Bai, H., and Tang, J.
\newblock Cascaded deep video deblurring using temporal sharpness prior.
\newblock In \emph{Proceedings of the IEEE/CVF Conference on Computer Vision and Pattern Recognition}, pp.\  3043--3051, 2020.

\bibitem[Park et~al.(2023)Park, Lee, and Chun]{park2023all}
Park, D., Lee, B.~H., and Chun, S.~Y.
\newblock All-in-one image restoration for unknown degradations using adaptive discriminative filters for specific degradations.
\newblock In \emph{2023 IEEE/CVF Conference on Computer Vision and Pattern Recognition (CVPR)}, pp.\  5815--5824. IEEE, 2023.

\bibitem[Peng et~al.(2022)Peng, Wang, Zhang, Wang, and Meng]{peng2022exact}
Peng, J., Wang, Y., Zhang, H., Wang, J., and Meng, D.
\newblock Exact decomposition of joint low rankness and local smoothness plus sparse matrices.
\newblock \emph{IEEE Transactions on Pattern Analysis and Machine Intelligence}, 2022.

\bibitem[Qin et~al.(2020)Qin, Wang, Bai, Xie, and Jia]{qin2020ffa}
Qin, X., Wang, Z., Bai, Y., Xie, X., and Jia, H.
\newblock Ffa-net: Feature fusion attention network for single image dehazing.
\newblock In \emph{Proceedings of the AAAI Conference on Artificial Intelligence}, volume~34, pp.\  11908--11915, 2020.

\bibitem[Sadek(2012)]{sadek2012svd}
Sadek, R.~A.
\newblock Svd based image processing applications: State of the art, contributions and research challenges.
\newblock \emph{International Journal of Advanced Computer Science and Applications}, 3\penalty0 (7), 2012.

\bibitem[Sedghi et~al.(2019)Sedghi, Gupta, and Long]{sedghi2018the}
Sedghi, H., Gupta, V., and Long, P.~M.
\newblock The singular values of convolutional layers.
\newblock In \emph{International Conference on Learning Representations}, 2019.

\bibitem[Shi et~al.(2016)Shi, Caballero, Husz{\'a}r, Totz, Aitken, Bishop, Rueckert, and Wang]{shi2016real}
Shi, W., Caballero, J., Husz{\'a}r, F., Totz, J., Aitken, A.~P., Bishop, R., Rueckert, D., and Wang, Z.
\newblock Real-time single image and video super-resolution using an efficient sub-pixel convolutional neural network.
\newblock In \emph{Proceedings of the IEEE conference on computer vision and pattern recognition}, 2016.

\bibitem[Song et~al.(2023)Song, He, Qian, and Du]{song2023vision}
Song, Y., He, Z., Qian, H., and Du, X.
\newblock Vision transformers for single image dehazing.
\newblock \emph{IEEE Transactions on Image Processing}, 32:\penalty0 1927--1941, 2023.

\bibitem[Sozykin et~al.()Sozykin, Chertkov, Schutski, PHAN, Cichocki, and Oseledets]{sozykinttopt}
Sozykin, K., Chertkov, A., Schutski, R., PHAN, A.-H., Cichocki, A., and Oseledets, I.
\newblock Ttopt: A maximum volume quantized tensor train-based optimization and its application to reinforcement learning.
\newblock In \emph{Advances in Neural Information Processing Systems}.

\bibitem[Stark(2013)]{stark2013image}
Stark, H.
\newblock \emph{Image recovery: theory and application}.
\newblock Elsevier, 2013.

\bibitem[Sun et~al.(2022)Sun, Chen, He, Wang, Feng, Han, Ding, Cheng, Li, and Wang]{sun2022singular}
Sun, Y., Chen, Q., He, X., Wang, J., Feng, H., Han, J., Ding, E., Cheng, J., Li, Z., and Wang, J.
\newblock Singular value fine-tuning: Few-shot segmentation requires few-parameters fine-tuning.
\newblock In \emph{Advances in Neural Information Processing Systems}, 2022.

\bibitem[Valanarasu et~al.(2022)Valanarasu, Yasarla, and Patel]{valanarasu2022transweather}
Valanarasu, J. M.~J., Yasarla, R., and Patel, V.~M.
\newblock Transweather: Transformer-based restoration of images degraded by adverse weather conditions.
\newblock In \emph{Proceedings of the IEEE/CVF Conference on Computer Vision and Pattern Recognition}, pp.\  2353--2363, 2022.

\bibitem[Wang et~al.(2020)Wang, Wang, Zhao, Chan, Xu, and Meng]{wang2020hyperspectral}
Wang, K., Wang, Y., Zhao, X.-L., Chan, J. C.-W., Xu, Z., and Meng, D.
\newblock Hyperspectral and multispectral image fusion via nonlocal low-rank tensor decomposition and spectral unmixing.
\newblock \emph{IEEE Transactions on Geoscience and Remote Sensing}, 58\penalty0 (11):\penalty0 7654--7671, 2020.

\bibitem[Wang et~al.(2022{\natexlab{a}})Wang, Li, Gou, Hu, and Peng]{wang2022relationship}
Wang, W., Li, B., Gou, Y., Hu, P., and Peng, X.
\newblock Relationship quantification of image degradations.
\newblock \emph{arXiv preprint arXiv:2212.04148}, 2022{\natexlab{a}}.

\bibitem[Wang et~al.(2017)Wang, Peng, Zhao, Leung, Zhao, and Meng]{wang2017hyperspectral}
Wang, Y., Peng, J., Zhao, Q., Leung, Y., Zhao, X.-L., and Meng, D.
\newblock Hyperspectral image restoration via total variation regularized low-rank tensor decomposition.
\newblock \emph{IEEE Journal of Selected Topics in Applied Earth Observations and Remote Sensing}, 11\penalty0 (4):\penalty0 1227--1243, 2017.

\bibitem[Wang et~al.(2022{\natexlab{b}})Wang, Yu, and Zhang]{wang2022zero}
Wang, Y., Yu, J., and Zhang, J.
\newblock Zero-shot image restoration using denoising diffusion null-space model.
\newblock In \emph{The Eleventh International Conference on Learning Representations}, 2022{\natexlab{b}}.

\bibitem[Wang et~al.(2022{\natexlab{c}})Wang, Cun, Bao, Zhou, Liu, and Li]{wang2022uformer}
Wang, Z., Cun, X., Bao, J., Zhou, W., Liu, J., and Li, H.
\newblock Uformer: A general u-shaped transformer for image restoration.
\newblock In \emph{Proceedings of the IEEE/CVF Conference on Computer Vision and Pattern Recognition}, pp.\  17683--17693, 2022{\natexlab{c}}.

\bibitem[Xiao et~al.(2022)Xiao, Fu, Liu, Wu, and Zha]{xiao2022image}
Xiao, J., Fu, X., Liu, A., Wu, F., and Zha, Z.-J.
\newblock Image de-raining transformer.
\newblock \emph{IEEE Transactions on Pattern Analysis and Machine Intelligence}, 2022.

\bibitem[Yang et~al.(2017)Yang, Tan, Feng, Liu, Guo, and Yan]{yang2017deep}
Yang, W., Tan, R.~T., Feng, J., Liu, J., Guo, Z., and Yan, S.
\newblock Deep joint rain detection and removal from a single image.
\newblock In \emph{Proceedings of the IEEE conference on computer vision and pattern recognition}, pp.\  1357--1366, 2017.

\bibitem[Zamir et~al.(2022{\natexlab{a}})Zamir, Arora, Khan, Munawar, Khan, Yang, and Shao]{zamir2022learning}
Zamir, S., Arora, A., Khan, S., Munawar, H., Khan, F., Yang, M., and Shao, L.
\newblock Learning enriched features for fast image restoration and enhancement.
\newblock \emph{IEEE Transactions on Pattern Analysis and Machine Intelligence}, 2022{\natexlab{a}}.

\bibitem[Zamir et~al.(2021)Zamir, Arora, Khan, Hayat, Khan, Yang, and Shao]{zamir2021multi}
Zamir, S.~W., Arora, A., Khan, S., Hayat, M., Khan, F.~S., Yang, M.-H., and Shao, L.
\newblock Multi-stage progressive image restoration.
\newblock In \emph{Proceedings of the IEEE/CVF conference on computer vision and pattern recognition}, pp.\  14821--14831, 2021.

\bibitem[Zamir et~al.(2022{\natexlab{b}})Zamir, Arora, Khan, Hayat, Khan, and Yang]{zamir2022restormer}
Zamir, S.~W., Arora, A., Khan, S., Hayat, M., Khan, F.~S., and Yang, M.-H.
\newblock Restormer: Efficient transformer for high-resolution image restoration.
\newblock In \emph{Proceedings of the IEEE/CVF Conference on Computer Vision and Pattern Recognition}, pp.\  5728--5739, 2022{\natexlab{b}}.

\bibitem[Zhang et~al.(2023)Zhang, Huang, Yao, Yang, Yu, Zhou, and Zhao]{zhang2023ingredient}
Zhang, J., Huang, J., Yao, M., Yang, Z., Yu, H., Zhou, M., and Zhao, F.
\newblock Ingredient-oriented multi-degradation learning for image restoration.
\newblock In \emph{Proceedings of the IEEE/CVF Conference on Computer Vision and Pattern Recognition}, pp.\  5825--5835, 2023.

\bibitem[Zhang et~al.(2015)Zhang, Zou, He, and Sun]{zhang2015accelerating}
Zhang, X., Zou, J., He, K., and Sun, J.
\newblock Accelerating very deep convolutional networks for classification and detection.
\newblock \emph{IEEE transactions on pattern analysis and machine intelligence}, 38\penalty0 (10):\penalty0 1943--1955, 2015.

\bibitem[Zhang et~al.(2018)Zhang, Li, Li, Wang, Zhong, and Fu]{zhang2018image}
Zhang, Y., Li, K., Li, K., Wang, L., Zhong, B., and Fu, Y.
\newblock Image super-resolution using very deep residual channel attention networks.
\newblock In \emph{Proceedings of the European conference on computer vision (ECCV)}, pp.\  286--301, 2018.

\bibitem[Zhang et~al.()Zhang, Xu, Liu, Yan, and Zuo]{zhangself}
Zhang, Z., Xu, R., Liu, M., Yan, Z., and Zuo, W.
\newblock Self-supervised image restoration with blurry and noisy pairs.
\newblock In \emph{Advances in Neural Information Processing Systems}.

\bibitem[Zhang et~al.(2022)Zhang, Zhao, Jin, Xu, Yang, and Yan]{zhang2022noiser}
Zhang, Z., Zhao, S., Jin, X., Xu, M., Yang, Y., and Yan, S.
\newblock Noiser: Noise is all you need for enhancing low-light images without task-related data.
\newblock \emph{arXiv preprint arXiv:2211.04700}, 2022.

\bibitem[Zheng et~al.(2021)Zheng, Ren, Cao, Hu, Wang, Song, and Jia]{zheng2021ultra}
Zheng, Z., Ren, W., Cao, X., Hu, X., Wang, T., Song, F., and Jia, X.
\newblock Ultra-high-definition image dehazing via multi-guided bilateral learning.
\newblock In \emph{2021 IEEE/CVF Conference on Computer Vision and Pattern Recognition (CVPR)}, pp.\  16180--16189. IEEE, 2021.

\bibitem[Zhou et~al.(2021{\natexlab{a}})Zhou, Xiao, Chang, Fu, Liu, Pan, and Zha]{zhou2021image}
Zhou, M., Xiao, J., Chang, Y., Fu, X., Liu, A., Pan, J., and Zha, Z.-J.
\newblock Image de-raining via continual learning.
\newblock In \emph{Proceedings of the IEEE/CVF Conference on Computer Vision and Pattern Recognition}, pp.\  4907--4916, 2021{\natexlab{a}}.

\bibitem[Zhou et~al.(2022{\natexlab{a}})Zhou, Yu, Huang, Zhao, Gu, Loy, Meng, and Li]{zhou2022deep}
Zhou, M., Yu, H., Huang, J., Zhao, F., Gu, J., Loy, C.~C., Meng, D., and Li, C.
\newblock Deep fourier up-sampling.
\newblock In \emph{Advances in Neural Information Processing Systems}, 2022{\natexlab{a}}.

\bibitem[Zhou et~al.(2023)Zhou, Huang, Guo, and Li]{zhou2023fourmer}
Zhou, M., Huang, J., Guo, C.-L., and Li, C.
\newblock Fourmer: An efficient global modeling paradigm for image restoration.
\newblock In \emph{International Conference on Machine Learning}, pp.\  42589--42601. PMLR, 2023.

\bibitem[Zhou et~al.(2022{\natexlab{b}})Zhou, Li, and Change~Loy]{zhou2022lednet}
Zhou, S., Li, C., and Change~Loy, C.
\newblock Lednet: Joint low-light enhancement and deblurring in the dark.
\newblock In \emph{Computer Vision--ECCV 2022: 17th European Conference, Tel Aviv, Israel, October 23--27, 2022, Proceedings, Part VI}, pp.\  573--589. Springer, 2022{\natexlab{b}}.

\bibitem[Zhou et~al.(2021{\natexlab{b}})Zhou, Ren, Emerton, Lim, and Large]{zhou2021imageudc}
Zhou, Y., Ren, D., Emerton, N., Lim, S., and Large, T.
\newblock Image restoration for under-display camera.
\newblock In \emph{Proceedings of the IEEE/CVF Conference on Computer Vision and Pattern Recognition}, pp.\  9179--9188, 2021{\natexlab{b}}.

\end{thebibliography}
\bibliographystyle{icml2024}

\newpage
\appendix

\makeatletter
\newcommand*\bigdot{\mathpalette\bigdot@{.5}}
\newcommand*\bigdot@[2]{\mathbin{\vcenter{\hbox{\scalebox{#2}{$\m@th#1\bullet$}}}}}
\makeatother

\onecolumn
\section{Proof of the Propositions}
\subsection{Proof of Theorem 1}
\label{sec: proof_1}
\begin{theorem}
For an arbitrary matrix $X \in \mathbb{R}^{h\times w}$ and random orthogonal matrices $P \in \mathbb{R}^{h\times h}, Q\in \mathbb{R}^{w\times w}$, the products of the $PX$, $XQ$, $PXQ$ have the same singular values with the matrix $X$.
\label{theorem:orth_appendix}
\end{theorem}

\textit{Proof.} According to the definition of Singular Value Decomposition (SVD), we can decompose matrix $X \in \mathbb{R}^{h\times w}$ into $USV^T$, where $U\in \mathbb{R}^{h\times h}$ and $V \in \mathbb{R}^{w\times w}$ indicate the orthogonal singular vector matrices, $S \in \mathbb{R}^{h\times w}$ indicates the diagonal singular value matrix. Thus $X^{'}$=$PXQ$=$PUSV^TQ$. Denotes $U^{'}$=$PU$ and $V^{'T}$=$V^TQ$, then $X^{'}$ can be decomposed into $U^{'}SV^{'T}$ if $U^{'}$ and $V^{'T}$ are orthogonal matrices.

\begin{align}
    U^{'-1}&=(PU)^{-1} = U^{-1}P^{-1} =U^{T}P^{T} = (PU)^{T} = U^{'T} \\
    (V^{'T})^{-1} &= (V^{T}Q)^{-1} = Q^{-1}(V^{T})^{-1} =Q^{T}V = (V^{T}Q)^{T} = V^{'}
\end{align}
Therefore, $U^{'}U^{'T}=I$ and $V^{'T}V^{'}=I$, where $I$ denotes the identity matrix, and $U^{'}$, $V^{'T}$ are orthogonal. $X^{'}$ and $X$ have the same singular values $S$, and the singular vectors of $X$ can be orthogonally transformed to $PU$, $Q^{T}V$. Correspondingly, it can be easily extended to the case of $PX$ and $XQ$.
{\hfill $\square$}

\subsection{Equivalence Proof of Equation 3 and IDFT}
\label{sec: proof_2}
\textbf{Proposition.} \textit{The signal formation principle in Equation 3 is equivalence to the definitive Inverse Discrete Fourier Transform (IDFT), where we restate the Equation 3 as following:}

\begin{equation}
\label{eq:2dfft}
    X = \frac{1}{hw}
\sum_{u=0}^{h-1}\sum_{v=0}^{w-1} G(u,v)e^{j2\pi(\frac{um}{h}+\frac{vn}{w})}, \ m\in \mathbb{R}^{h-1}, n \in\mathbb{R}^{w-1}.
\end{equation}

\textit{Proof.} For the two-dimensional signal $X \in \mathbb{R}^{h\times w}$, we can represent any point on it through IDFT. Supposing ($m$,$n$) and ($m^{'}$, $n^{'}$) are two random points on $X$, where $m$, $m^{'}$ $\in$ [$0$, $h$-$1$], $n$, $n^{'}$ $\in$ [$0$, $w$-$1$], and ($m$,$n$) $\neq$ ($m^{'}$, $n^{'}$), we have

\begin{align}\label{eq:fft1}
    X(m,n) &= \frac{1}{hw}
\sum_{u=0}^{h-1}\sum_{v=0}^{w-1} G(u,v)e^{j2\pi(\frac{um}{h}+\frac{vn}{w})}, 
\end{align}
\begin{align}\label{eq:fft2}
    X(m^{'},n^{'}) &= \frac{1}{hw}
\sum_{u=0}^{h-1}\sum_{v=0}^{w-1} G(u,v)e^{j2\pi(\frac{um^{'}}{h}+\frac{vn^{'}}{w})}. 
\end{align}
$X(m,n)$ represents the signal value at ($m$,$n$) position on $X$, and the same as $X(m^{'},n^{'})$. Thus, we can rewrite $X$ as

\begin{align}
   X=&\begin{bmatrix}
    X(0,0) &\cdots& X(0,w-1) \\
    \vdots &\ddots& \vdots \\
    X(h-1,0) & \cdots &X(h-1,w-1)
    \end{bmatrix} \nonumber \\
    =& \frac{1}{hw}
\sum_{u=0}^{h-1}\sum_{v=0}^{w-1}G(u,v) \; \bigdot 
\begin{bmatrix}
    e^{j2\pi(\frac{u0}{h}+\frac{v0}{w})} &\cdots& e^{j2\pi(\frac{u0}{h}+\frac{v(w-1)}{w})} \\
    \vdots &\ddots& \vdots \\
    e^{j2\pi(\frac{u(h-1)}{h}+\frac{v0}{w})} & \cdots &e^{j2\pi(\frac{u(h-1)}{h}+\frac{v(w-1)}{w})}
    \end{bmatrix} \nonumber \\
=& \frac{1}{hw}
\sum_{u=0}^{h-1}\sum_{v=0}^{w-1} G(u,v)e^{j2\pi(\frac{um}{h}+\frac{vn}{w})}, 
\end{align}
where $m$ $\in \mathbb{R}^{h-1}, n \in\mathbb{R}^{w-1}$. And the two-dimensional wave $e^{j2\pi(\frac{um}{h}+\frac{vn}{w})} \in \mathbb{R}^{h-1 \times w-1}$ denotes the base component. Therefore, the formation principle of \cref{eq:fft} is equivalent to the definitive IDFT, \ie \cref{eq:fft1,eq:fft2}.
{\hfill $\square$}

\section{Model details and Training Protocols}
\label{sec. model}
\begin{figure*}[h]
\small
\begin{center}
   \includegraphics[width=0.99\linewidth]{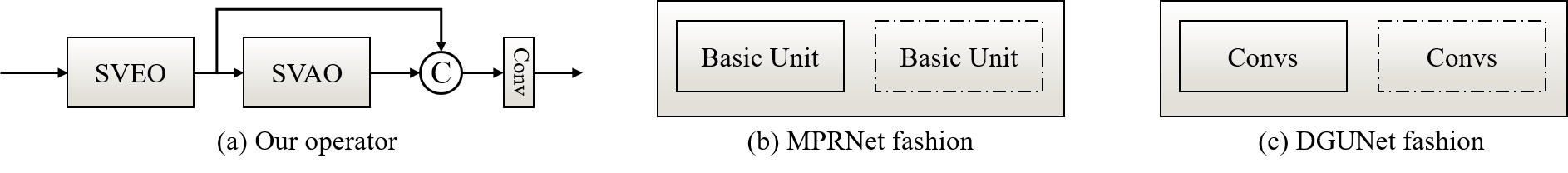}
\end{center}
\vspace{-1em}
    \caption{The strategy of model integration with DASL. (a) The working flow of our operator. (b) The basic building block of MPRNet fashion. (c) The basic building block of DGUNet fashion.}
\label{fig:DDLmodel}
\vspace{-0.2em}
\end{figure*}

We implement our DASL with integrated MPRNet~\cite{zamir2021multi}, DGUNet~\cite{mou2022deep}, and AirNet~\cite{li2022all} backbone to validate the effectiveness of the decomposed optimization.
All experiments are conducted using PyTorch, with model details and training protocols provided in \cref{tab:details}.
\cref{fig:DDLmodel} (a) presents the compound working flow of our operator. Note that the SVAO is only adopted in the bottleneck layer, as described in \cref{sec:overview}. 
We introduce how we embed our operator into the backbone network from a microscopic perspective. Sincerely, the most convenient way is to directly reform the basic block of the backbone network.
We present two fashions of the basic block of baseline in \cref{fig:DDLmodel} (b) and (c), where the MPRNet fashion is composed of two basic units, \eg, channel attention block (CAB)~\cite{zhang2018image}, and DGUNet is constructed by two vanilla activated convolutions.
We simply replace one of them (dashed line) with our operator to realize the DASL integration.
Note that AirNet shares the similar fashion as MPRNet.

\begin{table}[h]
    \centering
    \small
    \resizebox{\linewidth}{!}{
    \begin{tabular}{l |c c c}
    \toprule
         Configurations & MPRNet & DGUNet & AirNet \\
    \midrule
         optimizer & Adam & Adam & Adam\\
         base learning rate & 2e-4 & 1e-4 & 1e-3 \\
         learning rate schedular & Cosine decay & Cosine decay & Linear decay \\
         momentum of Adam & $\beta_1 = 0.9,\beta_2 = 0.999$ & $\beta_1 = 0.9,\beta_2 = 0.999$ & $\beta_1 = 0.9,\beta_2 = 0.99$ \\
         channel dimension & 80 & 80 & 256\\
         augmentation & RandomCropFlip & RandomCropFlip & RandomCropFlip\\
         \textit{num}. of replaced operator & 18 & 14 & 50\\
         basic block & channel attention block & activated convolution & degradation guided module\\
         optimization objective & CharbonnierLoss + EdgeLoss & CharbonnierLoss + EdgeLoss & L1Loss\\
    \bottomrule
    \end{tabular}}
    \caption{Model details and training protocols for DASL integrated baselines.}
    \label{tab:details}
\end{table}

\section{Trivial Ablations on operator design}
The ablation experiment on the choice of scale factor r in SVEO is provided in \cref{Tab:scale}. Note that the larger r will incur larger model size. We empirically set the scale ratio r in SVEO as 2. The working flow ablation of combined operator is provided in \cref{Tab:workflow}, and the compound fashion is preferred.
\begin{table*}[h]
    \begin{minipage}{0.46\linewidth}
        \centering
        \small
        \caption{Ablation experiments on the scale ratio $r$ in the SVEO (PSNR$\uparrow$).}
        \setlength{\tabcolsep}{0.8mm}{
        \begin{tabular}{cccccc|c}
            \toprule[1.3pt]
            Scale ratio $r$&Rain100L & SOTS& GoPro& BSD68 & LOL& \textbf{Avg.} \\
            \hline
            1&38.01 &31.55&26.88&25.84&20.93&28.64 \\ 
            2&38.02&31.57&26.91&25.82&20.96&28.66\\
            4&38.07&31.58&26.92&25.81&20.98&28.67\\
            \bottomrule[1.5pt]
        \end{tabular}}
        \label{Tab:scale}
    \end{minipage}
    \hspace{2mm}
    \begin{minipage}{0.52\linewidth}
        \centering
        \small
        \caption{Ablation experiments on the working flow of the combined operator (PSNR$\uparrow$).}
        \setlength{\tabcolsep}{0.7mm}{
        \begin{tabular}{cccccc|c}
            \toprule[1.3pt]
            Working flow&Rain100L & SOTS& GoPro& BSD68 & LOL& \textbf{Avg.} \\
            \hline
            cascaded&38.01 &31.55&26.88&25.84&20.87&28.63\\ 
            parallel&38.02&31.57&26.91&25.82&20.88&28.64\\
            cascaded + parallel&38.02&31.57&26.91&25.82&20.96&28.66\\
            \bottomrule[1.5pt]
        \end{tabular}}
        \label{Tab:workflow}
    \end{minipage}
\vspace{-1.5em}
\end{table*}

\section{Extension experiments for property validation}
\label{sec: exp}
In \cref{Tab:udc}, we provide the performance of DASL on real-world image restoration tasks, \ie, under-display camera (UDC) image enhancement. Typically, images captured under UDC system suffer from both blurring due to the spread point spread function, and lower light transmission rate. Compared to vanilla baseline models, DASL is capable of boosting the performance consistently. Note that the above experiments are performed on real-world UDC dataset~\cite{zhou2021imageudc} without any fine-tuning, validating the capability of the model for processing undesirable degradations. 
\cref{Tab:transformer} evaluates the potential of DASL integration on transformer-based image restoration backbone. Albeit the convolutional form of the developed decomposed operators, the supposed architecture incompatibility problem is not come to be an obstacle. Note that we replace the projection layer at the end of the attention mechanism with developed operators for transformer-based methods.

\begin{table*}[h]
\begin{minipage}{0.52\linewidth}
\centering
\small
\vspace{-1em}
\caption{Quantitative results of real-world image restoration tasks (under-display camera image enhancement) on TOLED and POLED datasets.}
\setlength{\tabcolsep}{1.2mm}{
\scriptsize
\begin{tabular}{l|ccc|ccc}
    \toprule[1.3pt]
     &\multicolumn{3}{c}{TOLED} & \multicolumn{3}{|c}{POLED}\\ 
    \textbf{Method} &PSNR$\uparrow$&SSIM$\uparrow$&LPIPS$\downarrow$&PSNR$\uparrow$&SSIM$\uparrow$&LPIPS$\downarrow$\\
    \midrule
    {\baseline MPRNet}&{\baseline24.69}&{\baseline0.707}&{\baseline0.347}&{\baseline8.34}&{\baseline0.365}&{\baseline0.798}\\
    {\baseline DGUNet}&{\baseline19.67}&{\baseline0.627}&{\baseline0.384}&{\baseline8.88}&{\baseline0.391}&{\baseline0.810}\\
    {\baseline AirNet}&{\baseline14.58}&{\baseline0.609}&{\baseline0.445}&{\baseline7.53}&{\baseline0.350}&{\baseline0.820}\\
    \midrule  DASL+MPRNet&25.65&0.733&0.326&8.95&0.392&0.788\\
    DASL+DGUNet&25.25&0.727&0.329&9.80&0.410&0.783\\
    DASL+AirNet&18.83&0.637&0.426&9.13&0.398&0.784\\
    \bottomrule[1.5pt]
\end{tabular}}
\label{Tab:udc}
\end{minipage}
\hspace{2mm}
\begin{minipage}{0.46\linewidth}
\centering
\small
\renewcommand\arraystretch {1.2}
\vspace{-1em}
\caption{Evaluating the generality of the DASL integration on transformer-based image restoration backbone among five common image restoration tasks (PSNR$\uparrow$).}
\setlength{\tabcolsep}{1.1mm}{
\scriptsize
\begin{tabular}{l|ccccc|c}
    \toprule[1.3pt]
    Methods&Rain100L & BSD68 & GoPro &SOTS & LOL & \textbf{Avg.} \\
    \hline
    {\baseline SwinIR}&{\baseline30.78} &{\baseline30.59}&{\baseline24.52}&{\baseline21.50}&{\baseline17.81}&{\baseline25.04} \\ 
    {\baseline Restormer}&{\baseline34.81}&{\baseline31.49}&{\baseline27.22}&{\baseline24.09}&{\baseline20.41}&{\baseline27.60}\\
    \hline
    DASL+SwinIR&33.53&30.84&25.72&24.10&20.36&26.91\\
    DASL+Restormer&35.79&31.67&27.35&25.90&21.39&28.42\\
    \bottomrule[1.5pt]
\end{tabular}}
\label{Tab:transformer}
\end{minipage}
\vspace{-1.5em}
\end{table*}

\section{Cultivating the SVD potential for image restoration.}
In fact, Singular Value Decomposition (SVD) has been widely applied for a range of image restoration tasks, such as image denoising, image compression, etc., attributing to the attractive rank properties~\cite{sadek2012svd} including \textit{truncated energy maximization} and \textit{orthogonal subspaces projection}.  
The former takes the fact that SVD provides the optima low rank approximation of the signal in terms of dominant energy preservation, which could greatly benefit the signal compression. The latter exploits the fact that the separate order of SVD-decomposed components are orthogonal, which inherently partition the signal into independent rank space, e.g., signal and noise space or range and null space for further manipulation, supporting the application of image denoising or even prevailing inverse problem solvers~\cite{wang2022zero}. 
Moreover, the SVD-based degradation analysis proposed in this work excavates another promising property of SVD from the vector-value perspective, which is essentially different from previous rank-based method.
Encouragingly, the above two perspectives have the potential to collaborate well and the separate order property is supposed to be incorporated into the DASL for sophisticated degradation relationship investigation in future works.
We note that the above two SVD perspectives have the opportunity to collaborate well and the separate order potential is supposed to be incorporated into the DASL for sophisticated degradation relationship investigation in future works.

\section{More decomposition ascribed analysis for degradations}
\begin{figure*}[t]
\small
\begin{center}
   \includegraphics[width=0.99\linewidth]{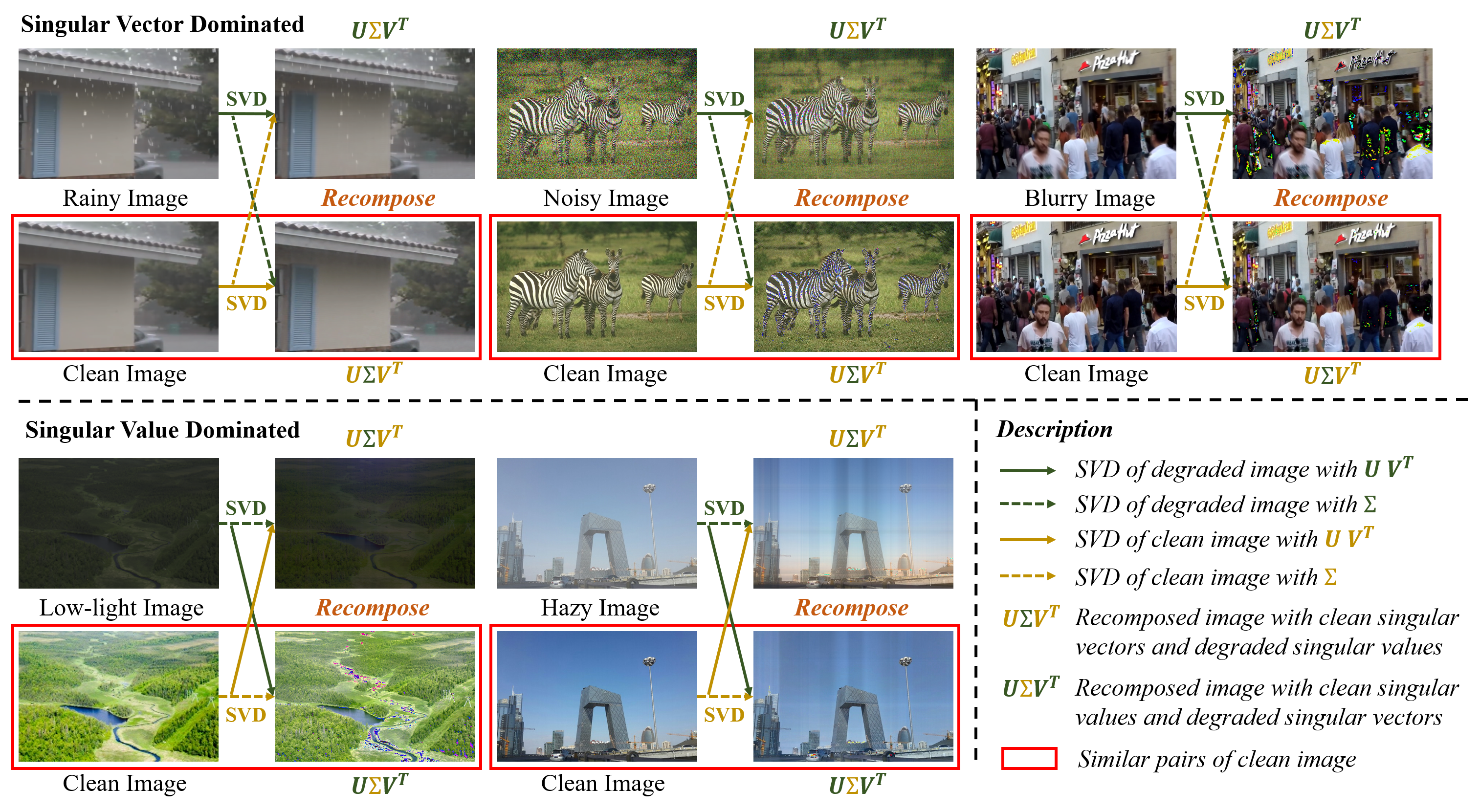}
\end{center}
\vspace{-1em}
    \caption{An illustration of the decomposition ascribed degradation analysis on various image restoration tasks through the lens of the singular value decomposition.}
\label{fig:supp1}
\end{figure*}

\begin{figure*}[t]
\small
\begin{center}
   \includegraphics[width=0.99\linewidth]{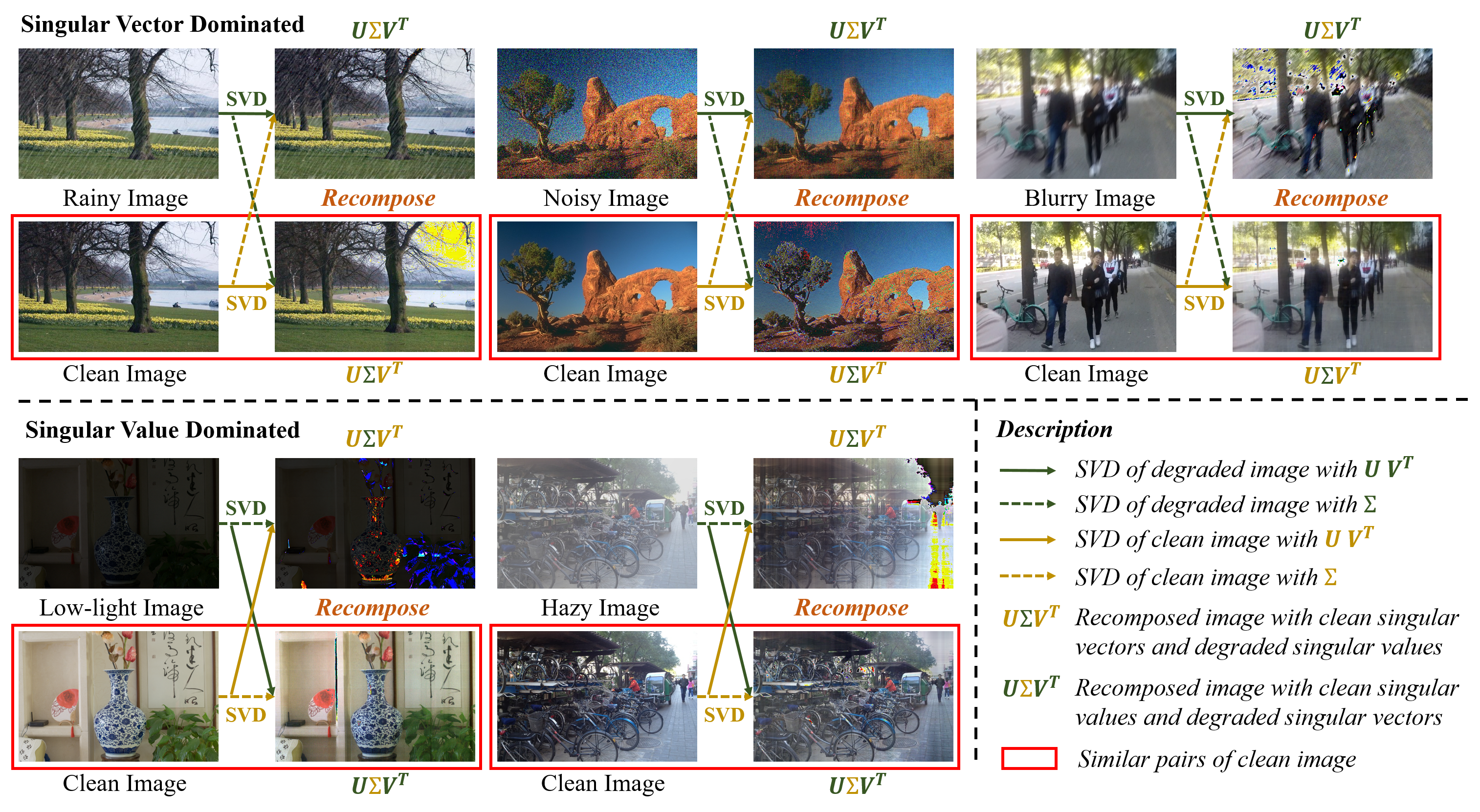}
\end{center}
\vspace{-1em}
    \caption{An illustration of the decomposition ascribed degradation analysis on various image restoration tasks through the lens of the singular value decomposition.}
\label{fig:supp2}
\end{figure*}

\begin{figure*}[t]
\small
\begin{center}
   \includegraphics[width=0.99\linewidth]{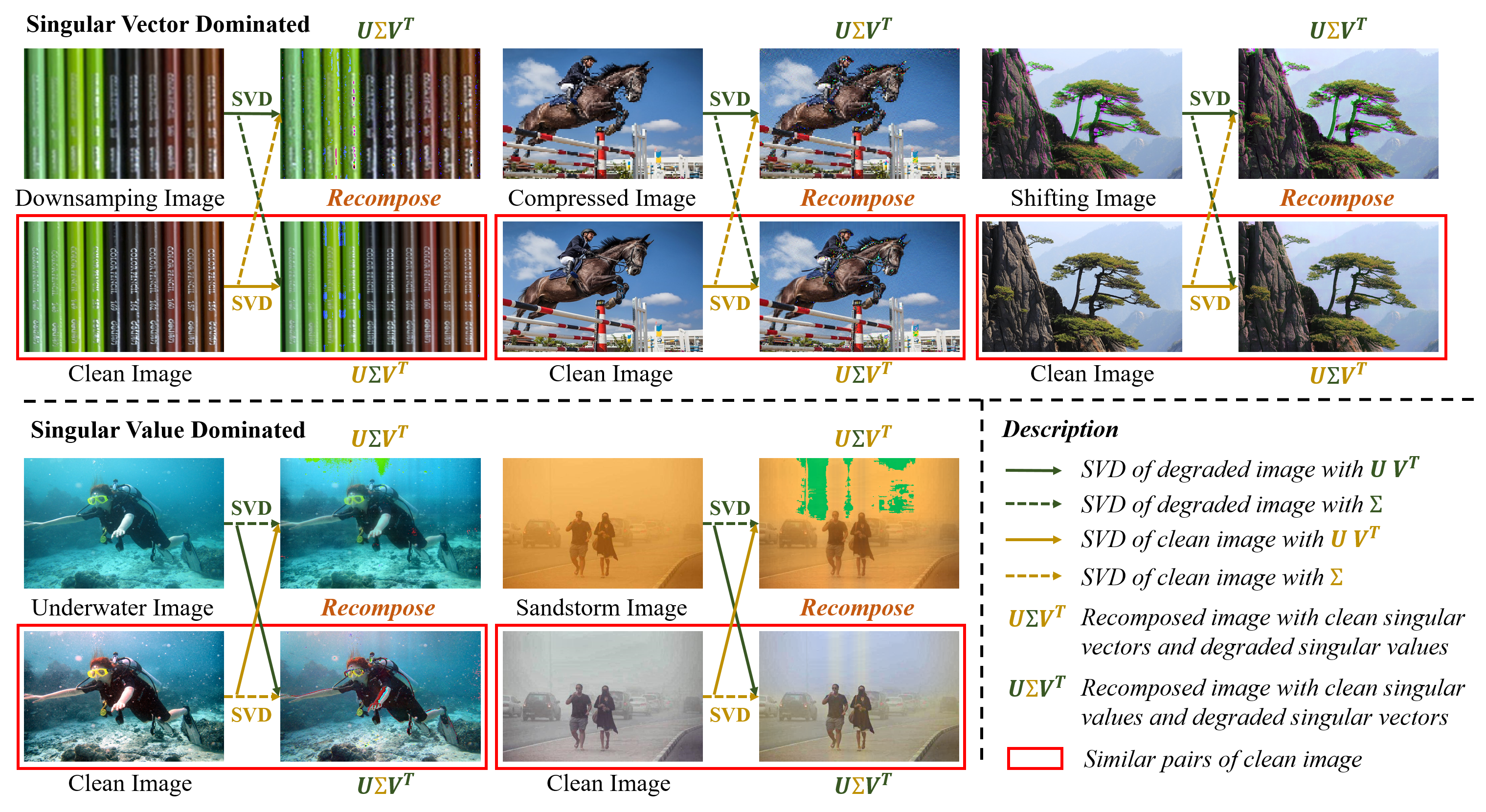}
\end{center}
\vspace{-1em}
    \caption{An illustration of the decomposition ascribed degradation analysis on various image restoration tasks through the lens of the singular value decomposition.}
\label{fig:supp3}
\end{figure*}

\begin{figure*}[h]
\small
\begin{center}
   \includegraphics[width=0.99\linewidth]{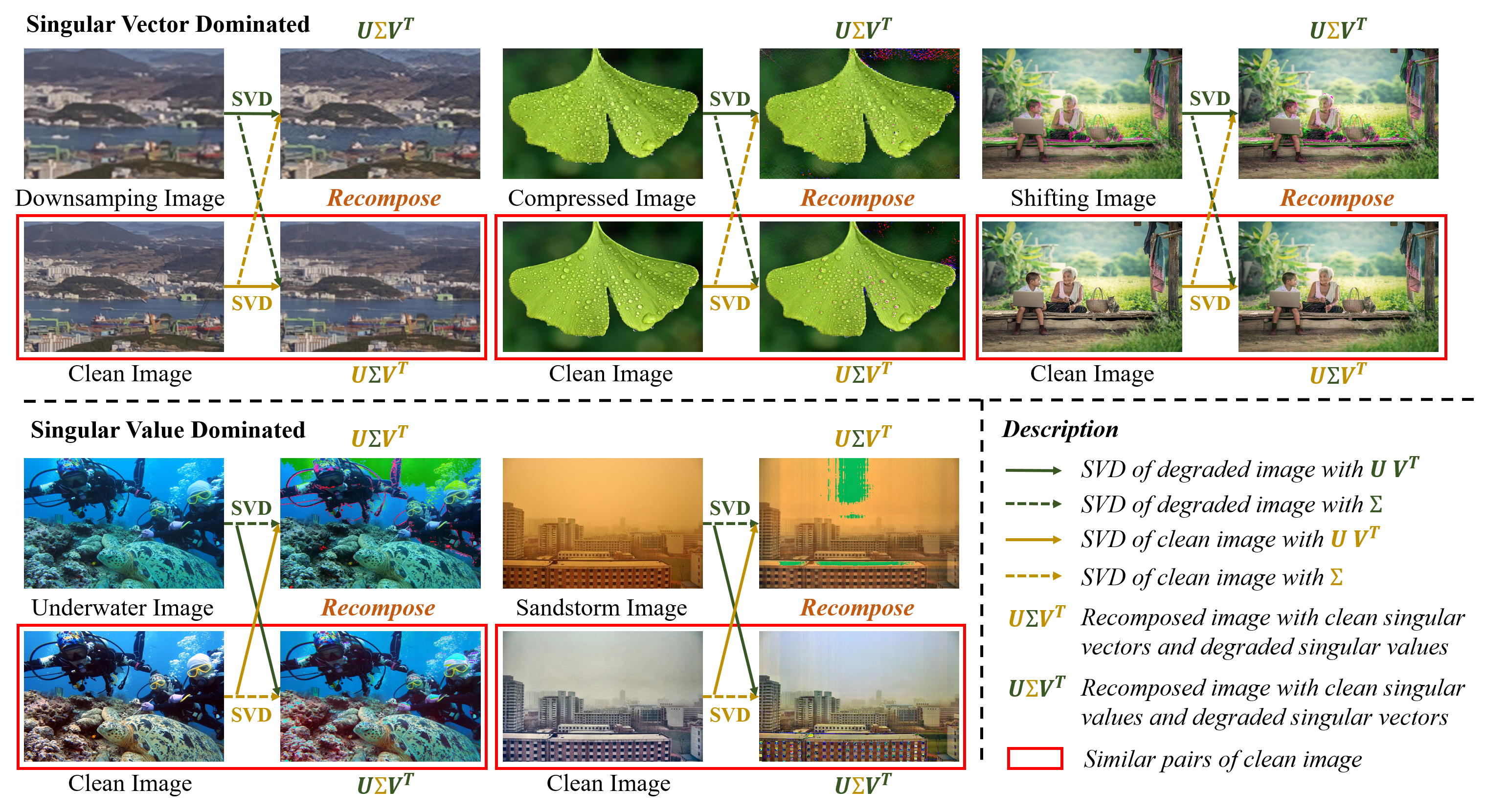}
\end{center}
\vspace{-1em}
    \caption{An illustration of the decomposition ascribed degradation analysis on various image restoration tasks through the lens of the singular value decomposition.}
\label{fig:supp4}
\end{figure*}

We provide more visual results of decomposition ascribed analysis for diverse degradations in \cref{fig:supp1,fig:supp2}, to further verify our observation that the decomposed singular vectors and singular values naturally undertake the different types of degradation information. 
In \cref{fig:supp3,fig:supp4}, we provide more degradation analysis to validate the generality of the proposed decomposition ascribed analysis, including \textit{downsampling, compression, color shifting, underwater enhancement, and sandstorm enhancement}. The former three types are ascribed into singular vector dominated degradations and the latter two types are ascribed into singular value dominated degradations.
Note that the aforementioned analysis also holds for real scene degradation. 

\section{Visual comparison results}
\label{sec:visual}
\begin{figure*}[htb]
\centering
    \begin{minipage}{0.186\textwidth}
        \centering
        \includegraphics[width=3.12cm]{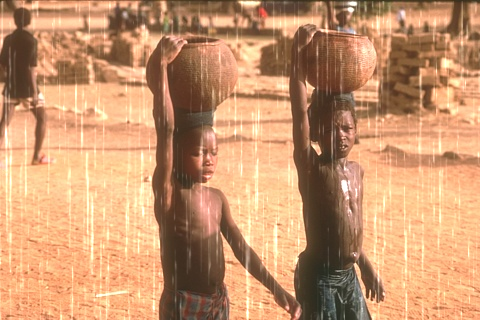}
        \subcaption*{Rainy Image}         
        \end{minipage}
    \begin{minipage}{0.186\textwidth}
		\centering
		\includegraphics[width=3.12cm]{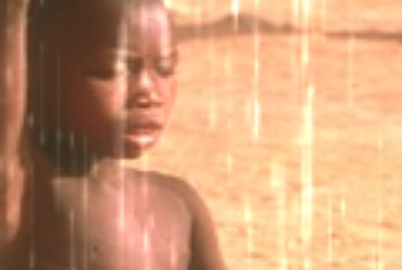}
		\subcaption*{Rainy}         
	\end{minipage}
    \begin{minipage}{0.186\textwidth}
		\centering
		\includegraphics[width=3.12cm]{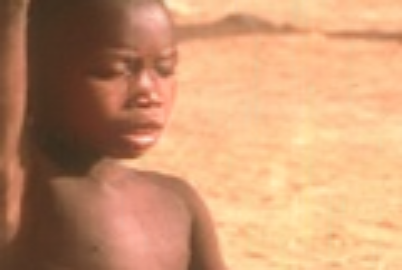}
		\subcaption*{NAFNet}         
	\end{minipage}
    \begin{minipage}{0.186\textwidth}
		\centering
		\includegraphics[width=3.12cm]{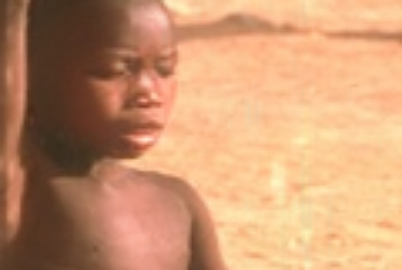}
		\subcaption*{HINet}         
	\end{minipage}
    \begin{minipage}{0.186\textwidth}
		\centering
		\includegraphics[width=3.12cm]{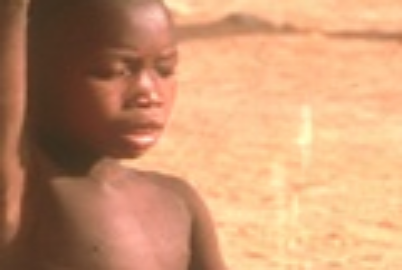}
		\subcaption*{MPRNet}         
	\end{minipage}

    \begin{minipage}{0.186\textwidth}
		\centering
		\includegraphics[width=3.12cm]{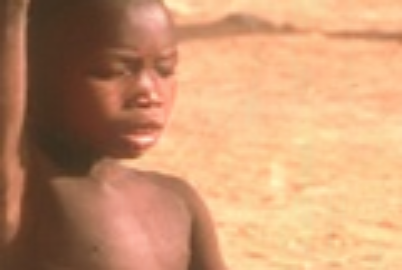}
		\subcaption*{DGUNet}         
	\end{minipage}
    \begin{minipage}{0.186\textwidth}
		\centering
		\includegraphics[width=3.12cm]{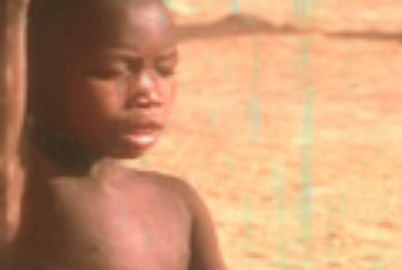}
		\subcaption*{MIRNetV2}         
	\end{minipage}
    \begin{minipage}{0.186\textwidth}
		\centering
		\includegraphics[width=3.12cm]{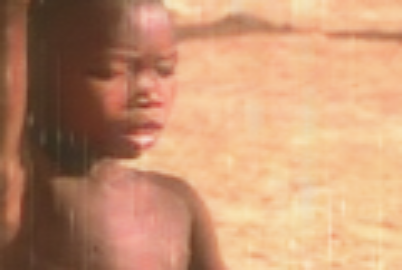}
		\subcaption*{Transweather}         
	\end{minipage}
    \begin{minipage}{0.186\textwidth}
		\centering
		\includegraphics[width=3.12cm]{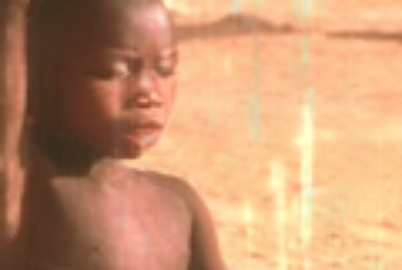}
		\subcaption*{SwinIR}         
	\end{minipage}
    \begin{minipage}{0.186\textwidth}
		\centering
		\includegraphics[width=3.12cm]{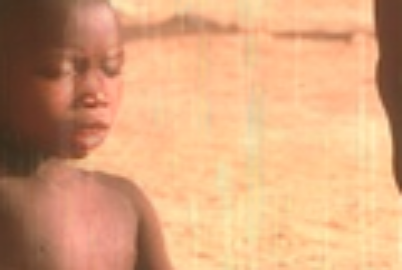}
		\subcaption*{TAPE}         
	\end{minipage}

    \begin{minipage}{0.186\textwidth}
		\centering
		\includegraphics[width=3.12cm]{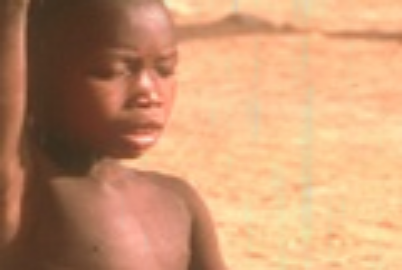}
		\subcaption*{IDR}      
        \end{minipage}
    \begin{minipage}{0.186\textwidth}
		\centering
		\includegraphics[width=3.12cm]{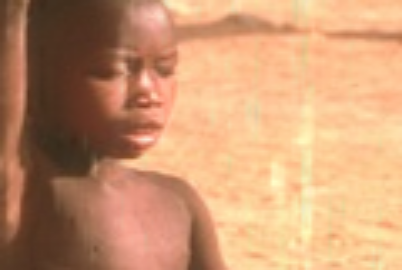}
		\subcaption*{Restormer}         
	\end{minipage}
    \begin{minipage}{0.186\textwidth}
		\centering
		\includegraphics[width=3.12cm]{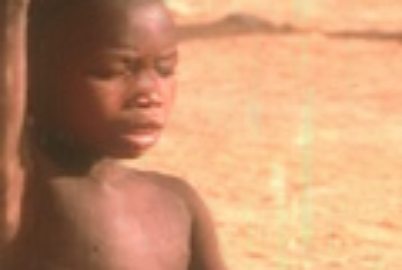}
		\subcaption*{AirNet}         
	\end{minipage}
    \begin{minipage}{0.186\textwidth}
		\centering
		\includegraphics[width=3.12cm]{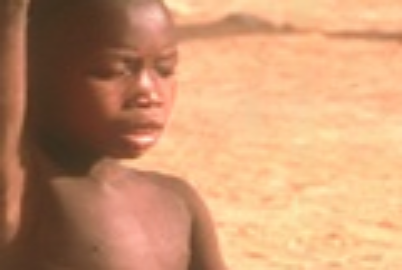}
		\subcaption*{DASL+MPRNet}         
	\end{minipage}
    \begin{minipage}{0.186\textwidth}
		\centering
		\includegraphics[width=3.12cm]{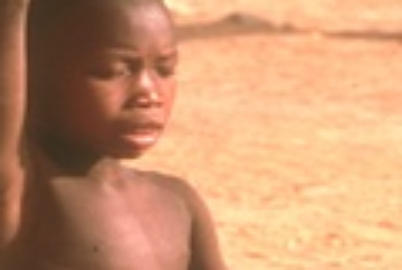}
		\subcaption*{Ground Truth}         
	\end{minipage}
\caption{Visual comparison with state-of-the-art methods on Rain100L dataset.} 
\vspace{-1em}
\label{fig:VC1}
\end{figure*}

\begin{figure*}[htb]
\centering
    \begin{minipage}{0.186\textwidth}
        \centering
        \includegraphics[width=3.12cm]{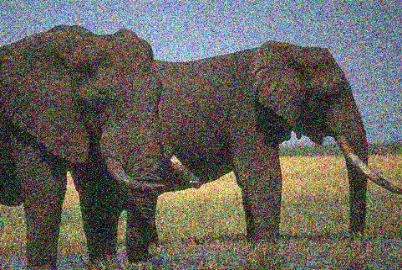}
        \subcaption*{Noisy Image}         
        \end{minipage}
    \begin{minipage}{0.186\textwidth}
		\centering
		\includegraphics[width=3.12cm]{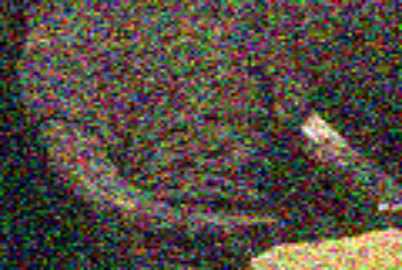}
		\subcaption*{Noisy}         
	\end{minipage}
    \begin{minipage}{0.186\textwidth}
		\centering
		\includegraphics[width=3.12cm]{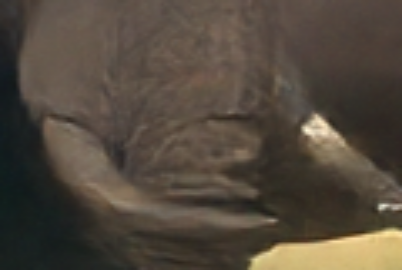}
		\subcaption*{NAFNet}         
	\end{minipage}
    \begin{minipage}{0.186\textwidth}
		\centering
		\includegraphics[width=3.12cm]{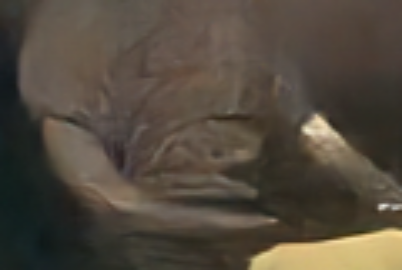}
		\subcaption*{HINet}         
	\end{minipage}
    \begin{minipage}{0.186\textwidth}
		\centering
		\includegraphics[width=3.12cm]{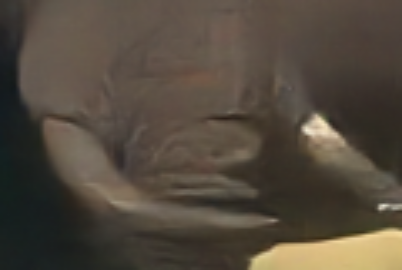}
		\subcaption*{MPRNet}         
	\end{minipage}

    \begin{minipage}{0.186\textwidth}
		\centering
		\includegraphics[width=3.12cm]{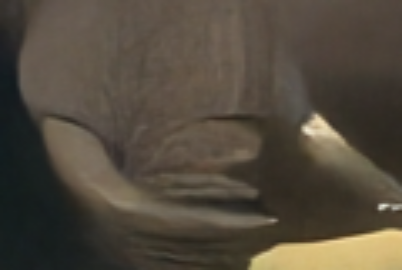}
		\subcaption*{DGUNet}         
	\end{minipage}
    \begin{minipage}{0.186\textwidth}
		\centering
		\includegraphics[width=3.12cm]{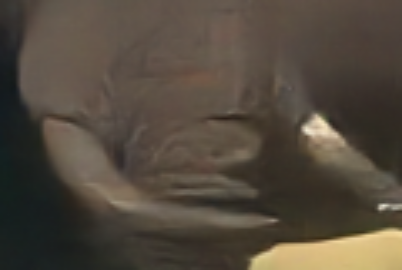}
		\subcaption*{MIRNetV2}         
	\end{minipage}
    \begin{minipage}{0.186\textwidth}
		\centering
		\includegraphics[width=3.12cm]{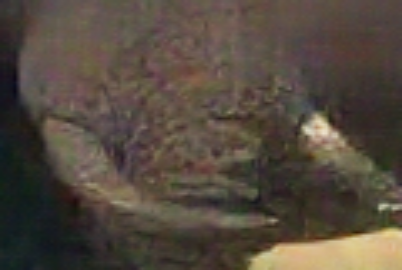}
		\subcaption*{Transweather}         
	\end{minipage}
    \begin{minipage}{0.186\textwidth}
		\centering
		\includegraphics[width=3.12cm]{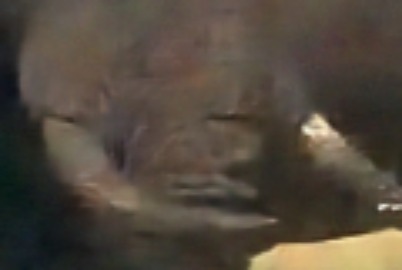}
		\subcaption*{SwinIR}         
	\end{minipage}
    \begin{minipage}{0.186\textwidth}
		\centering
		\includegraphics[width=3.12cm]{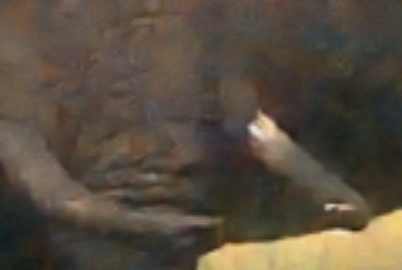}
		\subcaption*{TAPE}         
	\end{minipage}

    \begin{minipage}{0.186\textwidth}
		\centering
		\includegraphics[width=3.12cm]{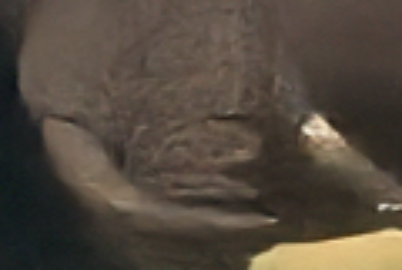}
		\subcaption*{IDR}         
	\end{minipage}
    \begin{minipage}{0.186\textwidth}
		\centering
		\includegraphics[width=3.12cm]{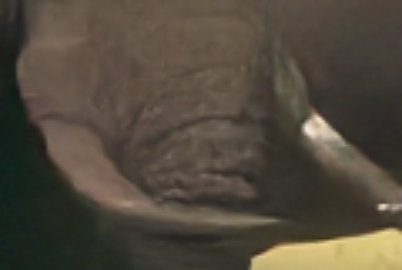}
		\subcaption*{Restormer}         
	\end{minipage}
    \begin{minipage}{0.186\textwidth}
		\centering
		\includegraphics[width=3.12cm]{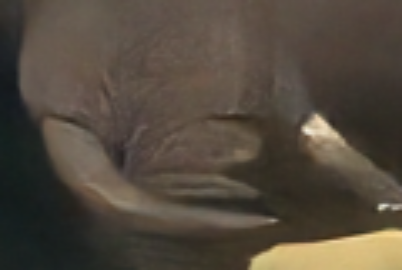}
		\subcaption*{AirNet}         
	\end{minipage}
    \begin{minipage}{0.186\textwidth}
		\centering
		\includegraphics[width=3.12cm]{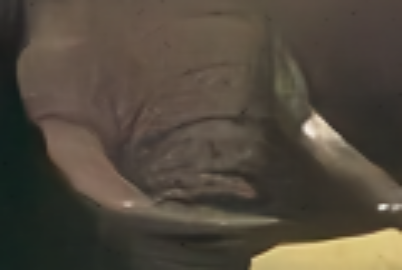}
		\subcaption*{DASL+MPRNet}         
	\end{minipage}
    \begin{minipage}{0.186\textwidth}
		\centering
		\includegraphics[width=3.12cm]{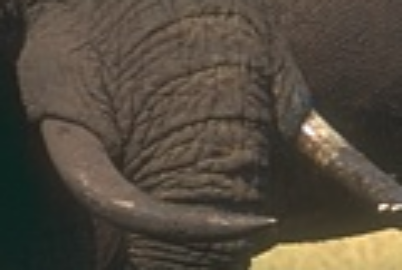}
		\subcaption*{Ground Truth}         
	\end{minipage}
\caption{Visual comparison with state-of-the-art methods on BSD68 dataset.} 
\vspace{-1em}
\label{fig:VC2}
\end{figure*}

\begin{figure*}[htb]
\centering
	\begin{minipage}{0.186\textwidth}
		\centering
		\includegraphics[width=3.12cm]{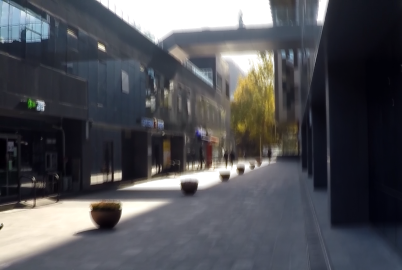}
		\subcaption*{Blurry Image}         
	\end{minipage}
    \begin{minipage}{0.186\textwidth}
		\centering
		\includegraphics[width=3.12cm]{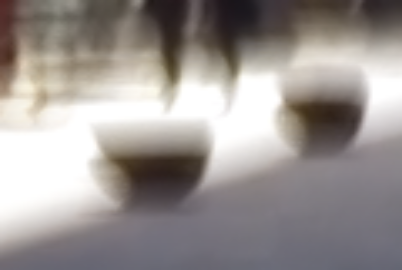}
		\subcaption*{Blurry}         
	\end{minipage}
    \begin{minipage}{0.186\textwidth}
		\centering
		\includegraphics[width=3.12cm]{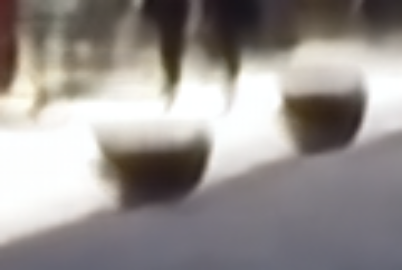}
		\subcaption*{NAFNet}         
	\end{minipage}
    \begin{minipage}{0.186\textwidth}
		\centering
		\includegraphics[width=3.12cm]{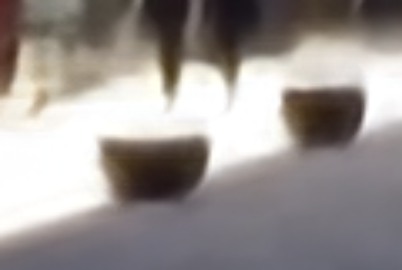}
		\subcaption*{HINet}         
	\end{minipage}
    \begin{minipage}{0.186\textwidth}
		\centering
		\includegraphics[width=3.12cm]{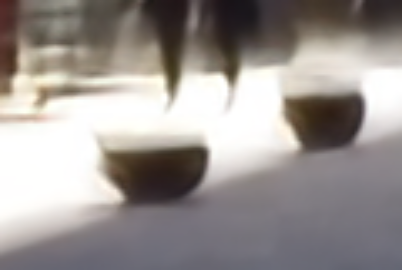}
		\subcaption*{MPRNet}         
	\end{minipage}

    \begin{minipage}{0.186\textwidth}
		\centering
		\includegraphics[width=3.12cm]{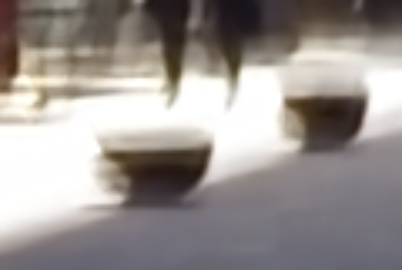}
		\subcaption*{DGUNet}         
	\end{minipage}
    \begin{minipage}{0.186\textwidth}
		\centering
		\includegraphics[width=3.12cm]{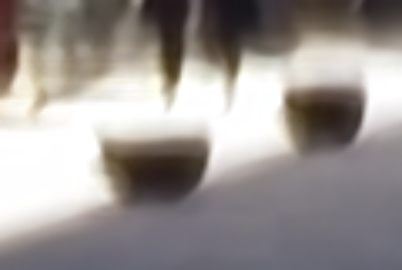}
		\subcaption*{MIRNetV2}         
	\end{minipage}
    \begin{minipage}{0.186\textwidth}
		\centering
		\includegraphics[width=3.12cm]{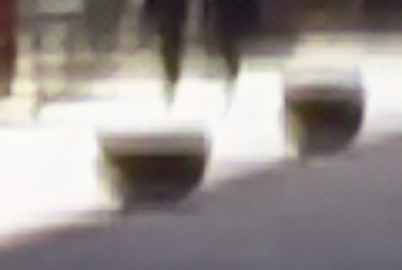}
		\subcaption*{Transweather}         
	\end{minipage}
    \begin{minipage}{0.186\textwidth}
		\centering
		\includegraphics[width=3.12cm]{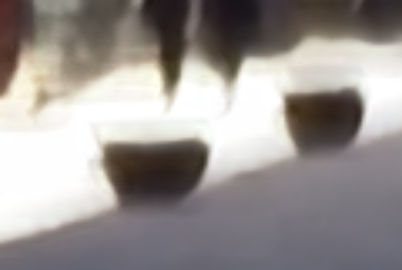}
		\subcaption*{SwinIR}         
	\end{minipage}
    \begin{minipage}{0.186\textwidth}
		\centering
		\includegraphics[width=3.12cm]{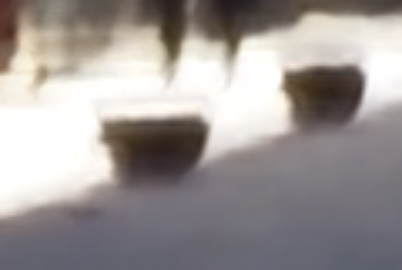}
		\subcaption*{TAPE}         
	\end{minipage}

    \begin{minipage}{0.186\textwidth}
		\centering
		\includegraphics[width=3.12cm]{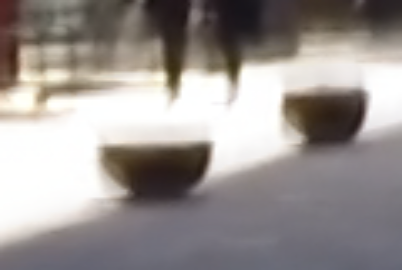}
		\subcaption*{IDR}         
	\end{minipage}
    \begin{minipage}{0.186\textwidth}
		\centering
		\includegraphics[width=3.12cm]{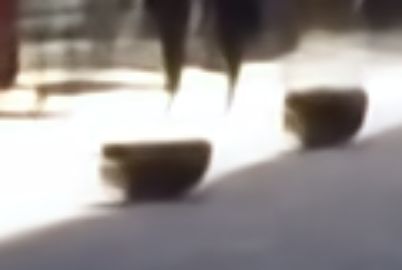}
		\subcaption*{Restormer}         
	\end{minipage}
    \begin{minipage}{0.186\textwidth}
		\centering
		\includegraphics[width=3.12cm]{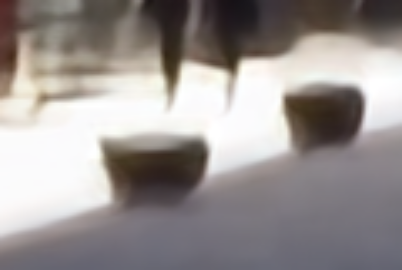}
		\subcaption*{AirNet}         
	\end{minipage}
    \begin{minipage}{0.186\textwidth}
		\centering
		\includegraphics[width=3.12cm]{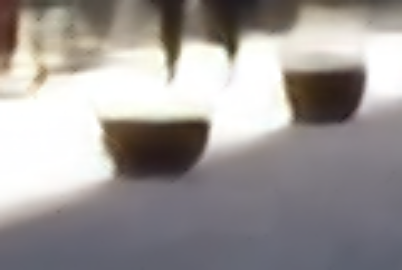}
		\subcaption*{DASL+MPRNet}         
	\end{minipage}
    \begin{minipage}{0.186\textwidth}
		\centering
		\includegraphics[width=3.12cm]{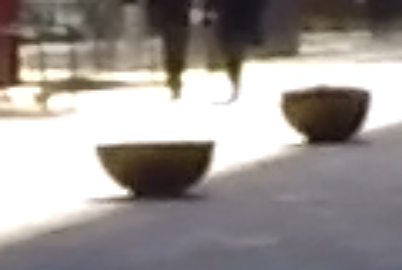}
		\subcaption*{Ground Truth}         
	\end{minipage}
\caption{Visual comparison with state-of-the-art methods on GoPro dataset.} 
\vspace{-1.5em}
\label{fig:VC3}
\end{figure*}

\begin{figure*}[htb]
\centering
	\begin{minipage}{0.186\textwidth}
		\centering
		\includegraphics[width=3.12cm]{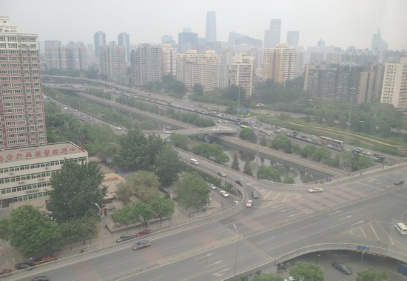}
		\subcaption*{Hazy Image}         
	\end{minipage}
    \begin{minipage}{0.186\textwidth}
		\centering
		\includegraphics[width=3.12cm]{baseline/hazy/ip.jpg}
		\subcaption*{Hazy}         
	\end{minipage}
    \begin{minipage}{0.186\textwidth}
		\centering
		\includegraphics[width=3.12cm]{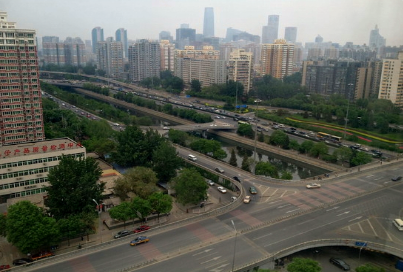}
		\subcaption*{NAFNet}         
	\end{minipage}
    \begin{minipage}{0.186\textwidth}
		\centering
		\includegraphics[width=3.12cm]{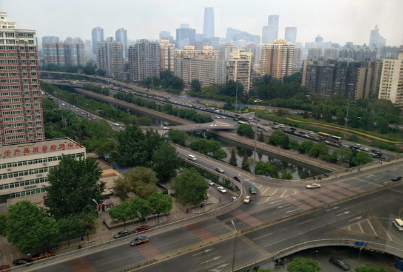}
		\subcaption*{HINet}         
	\end{minipage}
    \begin{minipage}{0.186\textwidth}
		\centering
		\includegraphics[width=3.12cm]{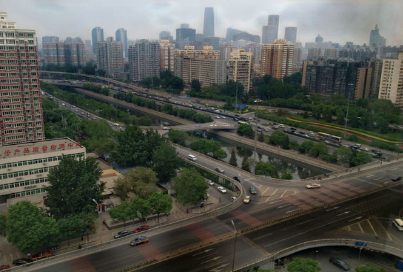}
		\subcaption*{MPRNet}         
	\end{minipage}

    \begin{minipage}{0.186\textwidth}
		\centering
		\includegraphics[width=3.12cm]{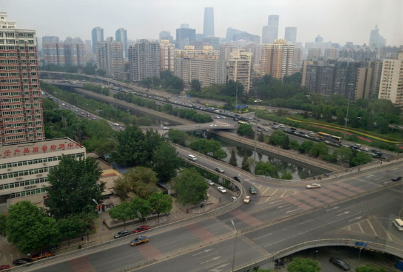}
		\subcaption*{DGUNet}         
	\end{minipage}
    \begin{minipage}{0.186\textwidth}
		\centering
		\includegraphics[width=3.12cm]{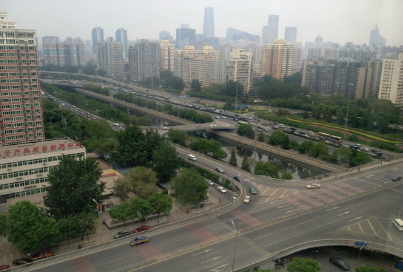}
		\subcaption*{MIRNetV2}         
	\end{minipage}
    \begin{minipage}{0.186\textwidth}
		\centering
		\includegraphics[width=3.12cm]{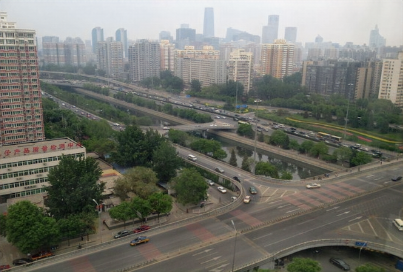}
		\subcaption*{Transweather}         
	\end{minipage}
    \begin{minipage}{0.186\textwidth}
		\centering
		\includegraphics[width=3.12cm]{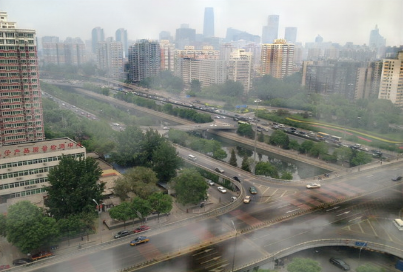}
		\subcaption*{SwinIR}         
	\end{minipage}
    \begin{minipage}{0.186\textwidth}
		\centering
		\includegraphics[width=3.12cm]{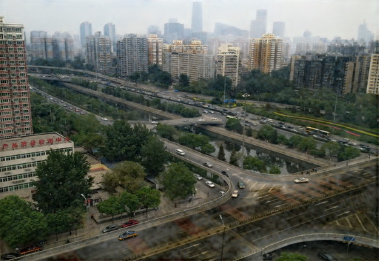}
		\subcaption*{TAPE}         
	\end{minipage}

    \begin{minipage}{0.186\textwidth}
		\centering
		\includegraphics[width=3.12cm]{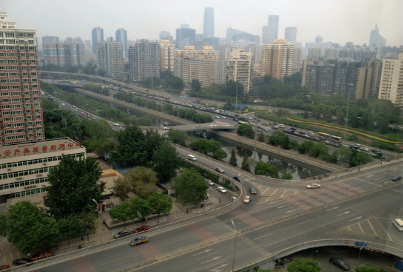}
		\subcaption*{IDR}         
	\end{minipage}
    \begin{minipage}{0.186\textwidth}
		\centering
		\includegraphics[width=3.12cm]{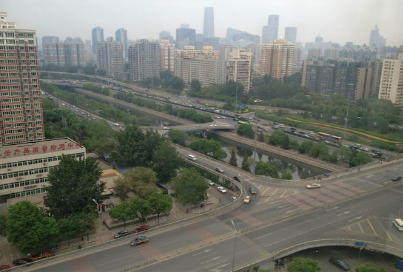}
		\subcaption*{Restormer}         
	\end{minipage}
    \begin{minipage}{0.186\textwidth}
		\centering
		\includegraphics[width=3.12cm]{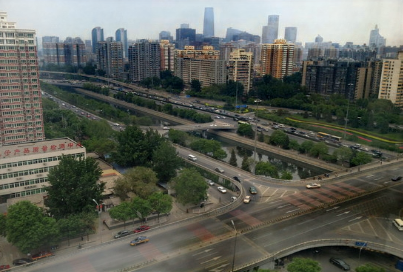}
		\subcaption*{AirNet}         
	\end{minipage}
    \begin{minipage}{0.186\textwidth}
		\centering
		\includegraphics[width=3.12cm]{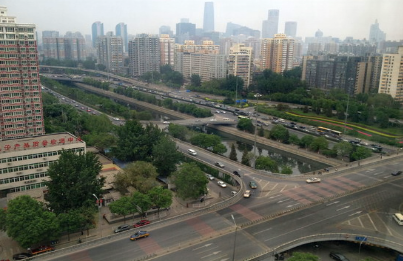}
		\subcaption*{DASL+MPRNet}         
	\end{minipage}
    \begin{minipage}{0.186\textwidth}
		\centering
		\includegraphics[width=3.12cm]{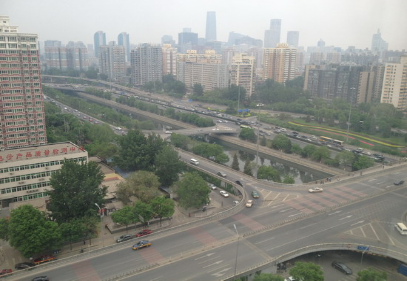}
		\subcaption*{Ground Truth}         
	\end{minipage}
\caption{Visual comparison with state-of-the-art methods on SOTS dataset.} 
\vspace{-1.2em}
\label{fig:VC4}
\end{figure*}

\begin{figure*}[htb]
\centering
	\begin{minipage}{0.186\textwidth}
		\centering
		\includegraphics[width=3.12cm]{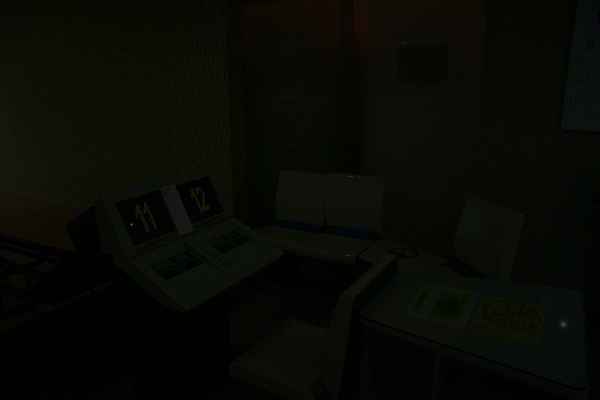}
		\subcaption*{Low-light Image}         
	\end{minipage}
    \begin{minipage}{0.186\textwidth}
		\centering
		\includegraphics[width=3.12cm]{baseline/low-light/ip.png}
		\subcaption*{Low-light}         
	\end{minipage}
    \begin{minipage}{0.186\textwidth}
		\centering
		\includegraphics[width=3.12cm]{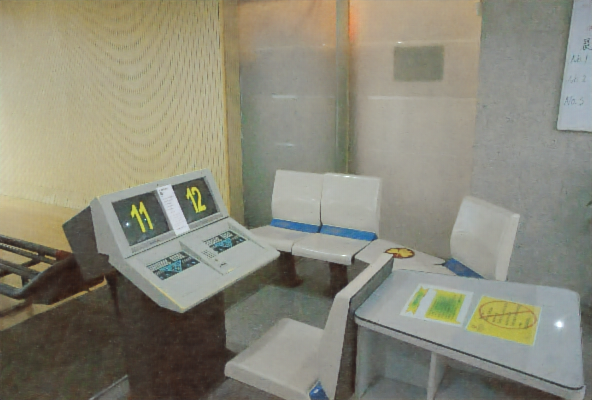}
		\subcaption*{NAFNet}         
	\end{minipage}
    \begin{minipage}{0.186\textwidth}
		\centering
		\includegraphics[width=3.12cm]{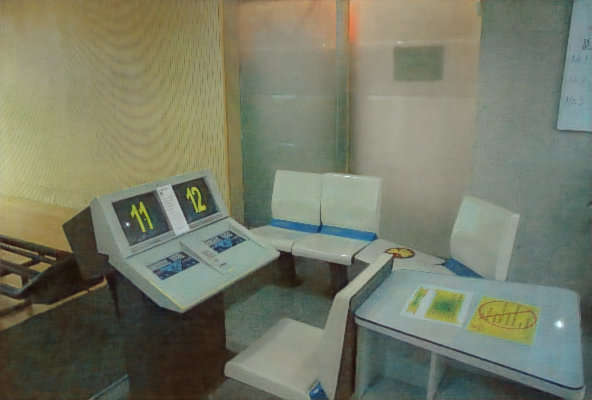}
		\subcaption*{HINet}         
	\end{minipage}
    \begin{minipage}{0.186\textwidth}
		\centering
		\includegraphics[width=3.12cm]{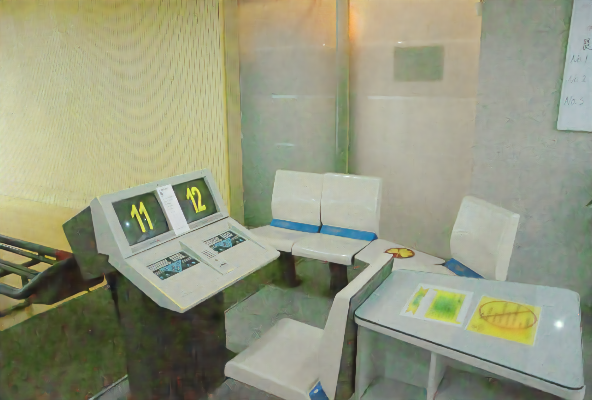}
		\subcaption*{MPRNet}         
	\end{minipage}

    \begin{minipage}{0.186\textwidth}
		\centering
		\includegraphics[width=3.12cm]{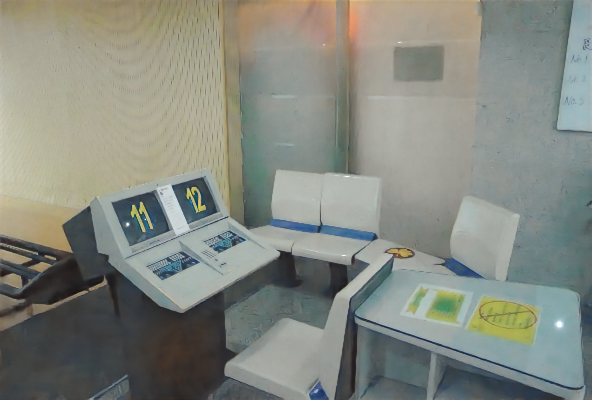}
		\subcaption*{DGUNet}         
	\end{minipage}
    \begin{minipage}{0.186\textwidth}
		\centering
		\includegraphics[width=3.12cm]{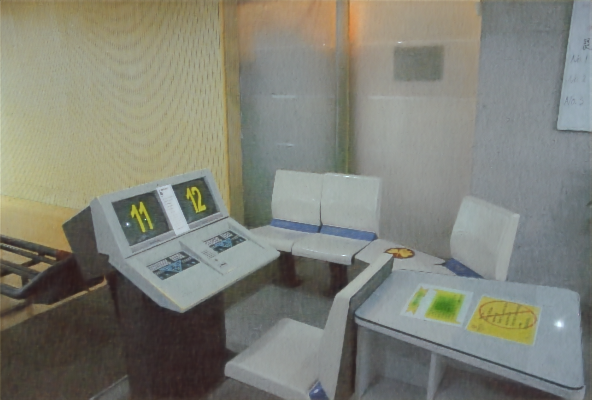}
		\subcaption*{MIRNetV2}         
	\end{minipage}
    \begin{minipage}{0.186\textwidth}
		\centering
		\includegraphics[width=3.12cm]{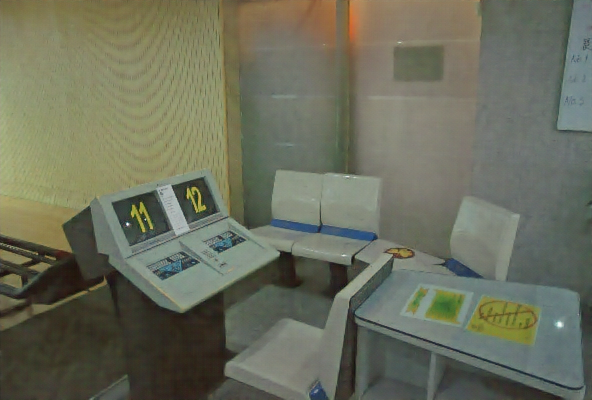}
		\subcaption*{Transweather}         
	\end{minipage}
    \begin{minipage}{0.186\textwidth}
		\centering
		\includegraphics[width=3.12cm]{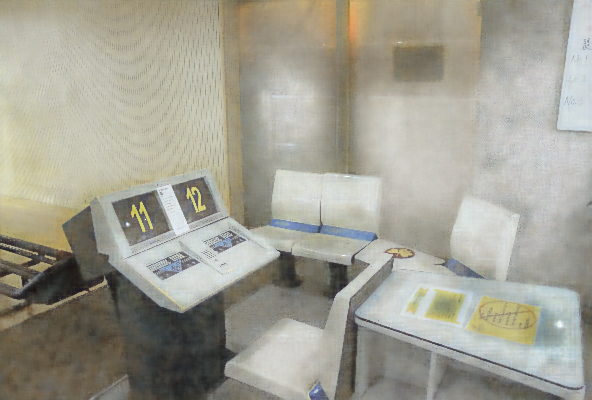}
		\subcaption*{SwinIR}         
	\end{minipage}
    \begin{minipage}{0.186\textwidth}
		\centering
		\includegraphics[width=3.12cm]{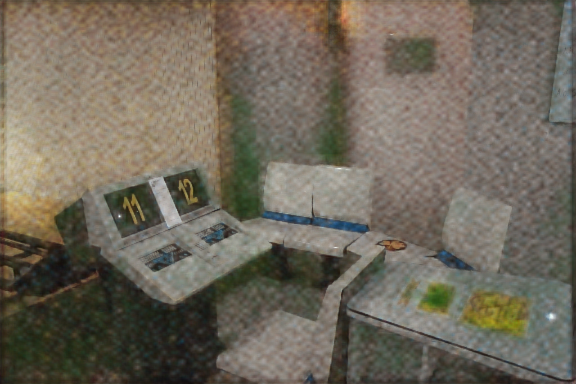}
		\subcaption*{TAPE}         
	\end{minipage}

    \begin{minipage}{0.186\textwidth}
		\centering
		\includegraphics[width=3.12cm]{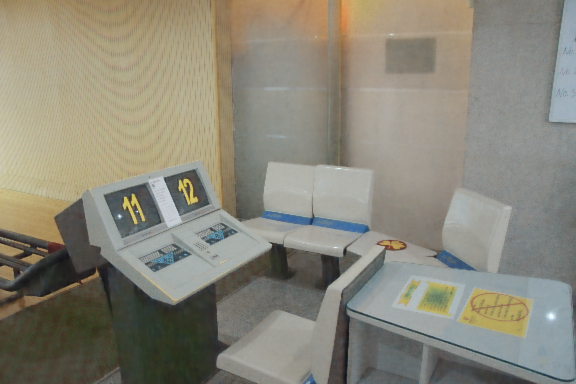}
		\subcaption*{IDR}         
	\end{minipage}
    \begin{minipage}{0.186\textwidth}
		\centering
		\includegraphics[width=3.12cm]{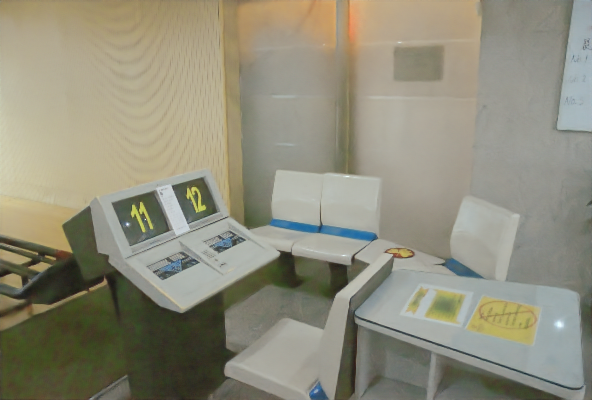}
		\subcaption*{Restormer}         
	\end{minipage}
    \begin{minipage}{0.186\textwidth}
		\centering
		\includegraphics[width=3.12cm]{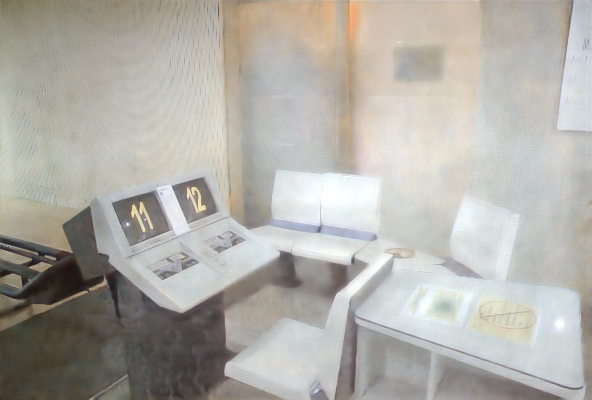}
		\subcaption*{AirNet}         
	\end{minipage}
    \begin{minipage}{0.186\textwidth}
		\centering
		\includegraphics[width=3.12cm]{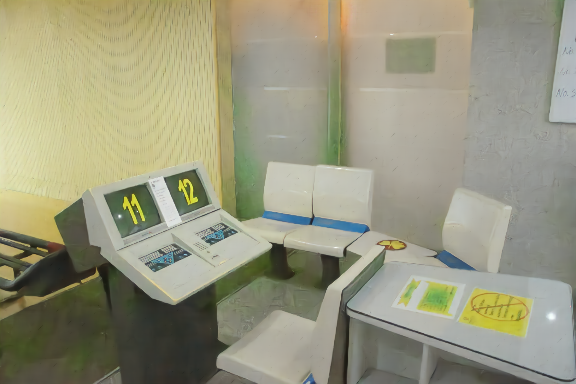}
		\subcaption*{DASL+MPRNet}         
	\end{minipage}
    \begin{minipage}{0.186\textwidth}
		\centering
		\includegraphics[width=3.12cm]{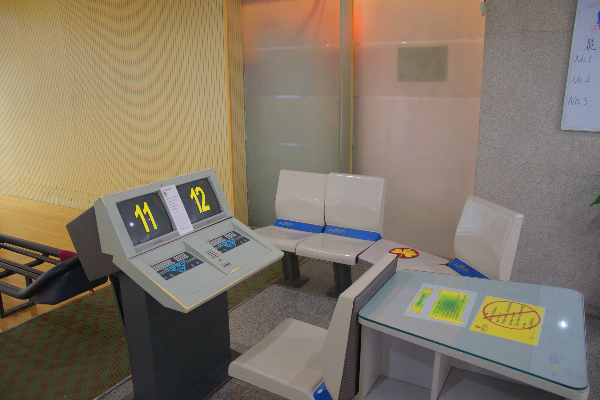}
		\subcaption*{Ground Truth}         
	\end{minipage}
\caption{Visual comparison with state-of-the-art methods on LOL dataset.} 
\vspace{-1.5em}
\label{fig:VC5}
\end{figure*}

We present the visual comparison results of the aforementioned image restoration tasks in \cref{fig:VC1,fig:VC2,fig:VC3,fig:VC4,fig:VC5}, including singular vector dominated degradations \textit{rain}, \textit{noise}, \textit{blur}, and singular value dominated degradations \textit{low-light}, \textit{haze}. It can be observed that our DASL exhibits superior visual recovery quality in both types of degradation, \ie, more precise content details in singular vector dominated degradations and more stable global recovery in singular value dominated degradations, compared to the integrated baseline method.

\end{document}